# Enhanced Conditional Generation of Double Perovskite by Knowledge-Guided Language Model Feedback

Inhyo Lee[1,†], Junhyeong Lee[1,†], Jongwon Park[1], KyungTae Lim[2,*], and Seunghwa Ryu[1,3*]

**Affiliations**

[1] Department of Mechanical Engineering, Korea Advanced Institute of Science and Technology (KAIST), 291 Daehak-ro, Yuseong-gu, Daejeon 34141, Republic of Korea

[2] Graduate School of Culture Technology, Korea Advanced Institute of Science and Technology (KAIST), 291 Daehak-ro, Yuseong-gu, Daejeon 34141, Republic of Korea

[3] KAIST InnoCORE PRISM-AI Center, Korea Advanced Institute of Science and Technology (KAIST), Daejeon, 34141, Republic of Korea

[*]Corresponding author: ryush@kaist.ac.kr, ktlim@kaist.ac.kr

**Abstract**

Double perovskites (DPs) are promising candidates for sustainable energy technologies due to their compositional tunability and compatibility with low-energy fabrication, yet their vast design space poses a major challenge for conditional materials discovery. This work introduces a multi-agent, text gradient–driven framework that performs DP composition generation under natural-language conditions by integrating three complementary feedback sources: LLM-based self-evaluation, DP-specific domain knowledge-informed feedback, and ML surrogate-based feedback. Analogous to how knowledge-informed machine learning improves the reliability of conventional data-driven models, our framework incorporates domain-informed text gradients to guide the generative process toward physically meaningful regions of the DP composition space. Systematic comparison of three incremental configurations, (i) pure LLM generation, (ii) LLM generation with LLM reasoning-based feedback, and (iii) LLM generation with domain knowledge–guided feedback, shows that iterative guidance from knowledge-informed gradients improves stability-condition satisfaction without additional training data, achieving over 98% compositional validity and up to ~54% stable or metastable candidates, surpassing both the LLM-only baseline (~43%) and prior GAN-based results (~27%). Analyses of ML-based gradients further reveal that they enhance performance in in-distribution (ID) regions but become unreliable in out-of-distribution (OOD) regimes. Overall, this work provides the first systematic analysis of multi-agent, knowledge-guided text gradients for DP discovery and establishes a generalizable blueprint for MAS-driven generative materials design aimed at advancing sustainable technologies.

## 1. Introduction

The pursuit of sustainable materials has emerged as a central challenge in addressing the global energy and environmental crisis.[1–3] Among various classes of materials, double perovskites (DPs) have emerged as one of the most promising sustainable materials owing to their highly tunable optoelectronic properties, broad compositional versatility, and compatibility with low-energy, solution-based fabrication. [4–6] It enables low-temperature processing, which could contribute to reduced energy consumption during fabrication.[7,8] Furthermore, the compositional flexibility of DPs allows for the replacement of toxic or scarce elements with abundant, environmentally benign alternatives, enhancing both material sustainability and long-term scalability.[5,9,10] Their superior performance in solar cells, photocatalysts, batteries, and thermoelectric converters further highlights their potential to underpin a carbon-neutral energy economy.[11–13]

However, this very compositional diversity also defines a vast and complex design space, often encompassing hundreds of thousands of possible compounds, making conventional trial-and-error or first-principles exploration computationally prohibitive. To overcome this combinatorial complexity, data-driven approaches have been increasingly employed to discover perovskite materials with desired properties.[14–20] For instance, semi-supervised variational autoencoders (SS-VAEs) have been combined with genetic algorithms to identify thermodynamically stable DPs, while hybrid SS-VAE–GAN frameworks have been developed to generate structures following predefined geometry.[21,22]

Although data-driven generative models, including VAE, Generative Adversarial Network (GAN), and diffusion-based frameworks have shown potential in navigating the vast compositional space, they still face two major limitations. First, many generative models fail to incorporate explicit physical constraints during optimization in the latent space.[21,23–25] As a result, the decoded structures often exhibit atomic overlaps, or low-symmetry configurations, where physical plausibility is not guaranteed. Second, their performance remains highly dependent on the training data distribution, which confines generation to known chemical domains and limits the exploration of unseen compositions.[26–29]

In contrast, large language models (LLMs) offer a potential solution to the first limitation. By generating sequences through next-token prediction, LLMs implicitly capture the underlying chemical grammar of elements and bonds. When trained on large corpora of chemical strings, this grammar emerges implicitly as a set of statistical regularities over valid element tokens, bonding patterns, and token order, allowing them to preserve the syntactic validity of string-based representation such as compositions or SMILES notations. Moreover, since crystal structures can also be expressed in string formats (e.g., crystallographic information files), LLMs are inherently well-suited to generating structurally valid configurations as well. Recent studies have confirmed their higher fidelity compared with VAEs, GANs, or diffusion models, and LLMs are now increasingly used in materials generation frameworks.[30–32]

For the second limitation, although LLMs are often perceived as capable of extrapolating beyond their training distribution, they remain fundamentally data-driven generative models, and their ability to perform true extrapolation is inherently limited. Consequently, to introduce patterns that the model has not previously learned, several recent studies have enhanced performance in materials and chemical domains by fine-tuning LLMs with domain-specific datasets, thereby enabling more accurate and contextually grounded reasoning within the target chemical space.[31,33–36] For example, recent work has shown that fine-tuning LLMs on text-encoded atomistic representations markedly improves their ability to generate physically plausible and thermodynamically meaningful crystal structures, outperforming conventional generative models.[37] In addition, another study adopted an iterative fine-tuning strategy to leverage the model's limited near-boundary extrapolative behavior, gradually improving its generative performance across iterations.[38] However, these fine-tuning-based approaches generally require substantial computational resources, which restrict their scalability and broader applicability.

However, unlike conventional generative models, LLMs offer an alternative mechanism for improving predictive and generative performance without modifying their vast number of parameters. Leveraging their massive training corpora and high-dimensional latent manifold, LLMs exhibit emergent in-context learning (ICL) capabilities, enabling them to adapt to new tasks solely through examples or structured prompts, without any weight updates.[39–42] As a result, model performance can often be enhanced simply by enriching the prompt with informative and task-relevant demonstrations.[34,43–45] Furthermore, agent-based and retrieval-augmented generation (RAG) strategies have been proposed to facilitate more effective prompt construction by integrating external tools, databases, and domain-specific knowledge.[46–51]

Building on these developments, recent studies have explored the use of multi-agent LLM frameworks to develop more scalable and reliable systems. Such frameworks enable coordinated reasoning, tool-calling, and interaction among multiple AI agents, thereby addressing the inherent single-agent systems.[52–54] By allowing specialized agents to collaborate and exchange intermediate reasoning results, multi-agent systems (MAS) have shown strong effectiveness in tackling complex, open-ended problems that require iterative exploration, critique, and self-evaluation. For example, in the materials domain, representative work such as ChatMOF employed a multi-agent approach for the automated discovery of metal-organic frameworks, in which multiple LLM agents collaboratively generated, assessed, and refined candidate structures.[55] These early demonstrations highlight the promise of MAS in extending LLM-driven reasoning to more intricate and high-dimensional materials design challenges. Another representative example is ChemCrow, which integrates an LLM with specialized chemistry tools to autonomously perform complex synthesis and materials design tasks, effectively bridging computational and experimental workflows.[56] More broadly, these and other emerging MAS-based approaches highlight the expanding role of agentic LLM frameworks in tackling complex materials design challenges.[57–61]

Within this paradigm, several studies have proposed prompt optimization or input adaptation strategies for MAS to improve accuracy of materials design.[62–64] Among them, one of the representative works, inspired by the backpropagation process in neural networks, proposed concept of a text gradient in LLMs. This framework interprets language-model feedback as pseudo-gradients that iteratively refine prompt or textual components in the MAS.[62] This work has applied this feedback-based methodology to molecular design based on the LLM self-evaluation and feedback, so-called text gradient. Likewise, another paper LLMatDesign integrated ML surrogates into a language-model feedback loop for hypothesis-driven materials generation for the materials design. However, in these studies, the feedback originates solely from LLM self-evaluation and hypotheses from ML surrogate prediction, without incorporating materials-specific domain knowledge or validating the reliability of surrogate models in out-of-domain (OOD). Given that LLMs are prone to hallucination or confusing beliefs with factual knowledge [65] and Machine learning (ML) surrogates prediction reliability can be deteriorated in OOD, compared to in-domain (ID),[66] such self-referential feedback loops or purely ML-driven hypotheses may provide unreliable guidance, as the feedback becomes ineffective when operating in OOD parts of the materials design space.

To overcome these limitations, this study introduces an optimization framework within a MAS designed to perform composition-generation tasks for DPs under given input conditions, incorporating concept of knowledge-informed text gradients, which argument language-model feedback with DP-specific domain knowledge and ML-based inference results, and analyzed their effectiveness on the composition generation progress by MAS. Analogous to how physics-informed machine learning incorporate physical constrains into the loss function to improve data efficiency,[67,68] our approach enriches text gradient feedback within DP-specific domain knowledge and ML-predicted heuristics. This integration is intended to enable more structured and knowledge-grounded exploration of the DP compositional space within the MAS, guiding the agentic workflow toward physically plausible reasoning paths. Furthermore, the proposed framework provides a means to systematically examine how the usefulness of ML-derived text gradients may vary across ID and OOD regions.

To strategically analyze these effects, **Section 2.1** defines three case studies: (1) pure LLM generation, (2) optimization with LLM feedback, and (3) optimization with DP-specific domain knowledge-informed feedback. **Section 2.2** describes the construction of test queries, followed by comparative analyses of generated compositions in **Section 2.3**. **Section 2.4** extends these cases by adding ML-driven text gradients and evaluates their impact on conditional generation performance across in-domain (ID) and out-of-domain (OOD) spaces. Subsequently, **Section 2.5** outlines the limitations and future directions of this work, and **Section 3** presents the overall conclusions. Finally, the **Methods** section details the POSCAR formulation, density functional theory (DFT) calculations, MAS architecture, and validation procedure.

## 2. Results & Discussion

To clarify the problem setting, our task can be viewed as guiding an LLM agent to propose double perovskite (DP) compositions that align well with a given set of input conditions, where better alignment corresponds to higher fitness (or equivalently, lower loss), within the DP compositional space (**Figure 1a**). In other words, the objective is to optimize the input composition $x$ such that the resulting textual fitness (or loss) is improved under the specified condition. Prior studies have approached this optimization problem by navigating the compositional landscape using LLM-based reasoning, relying on text gradients to iteratively refine candidate compositions (arrow in green color).[62] Meanwhile, another framework explored a complementary strategy, generating hypotheses from ML-predicted properties to guide the search process (arrow in red color).[57] Building on these perspectives, the present study investigates how incorporating DP-specific domain knowledge into the text gradient affects the optimization trajectory (arrow in blue color), and further examines how ML-driven gradients influence search behavior in both in ID and OOD regions.

To realize the above concept, as illustrated in **Figure 1b**, we first constructed a MAS by connecting a proposal agent with multiple evaluation agents, enabling the system to generate a candidate composition and subsequently assess it from different evaluative perspectives. In this system, three evaluation agents independently produce natural language evaluation output that we interpret as text-based loss, qualitative loss signals expressed in text that metaphorically play the role of an optimization objective. Specifically, the LLM-based evaluation agent relies solely on pretrained knowledge and yields $L_{LLM}$, the knowledge-based evaluation agent incorporates expert-curated domain rules and outputs $L_{DK}$, and the ML-based evaluation agent uses user-provided surrogate models to approximate a target property and produces $L_{ML}$. Building on this structure, we incorporated a reasoning-based gradient engine that reads each text-based loss and reasons about how the proposed composition should be improved to better satisfy the given condition, with the resulting gradient metaphorically denoting the direction of improvement in the optimization process. For each loss type, the gradient engine infers the direction of improvement and generates a corresponding text gradient, namely, the LLM-based gradient $\frac{\partial L_{LLM}}{\partial x}$, the knowledge-informed gradient $\frac{\partial L_{DK}}{\partial x}$, and the ML-based gradient $\frac{\partial L_{ML}}{\partial x}$, each describing the modification strategy for the next iteration. These gradients are then fed back into the proposal agent, closing the loop and enabling iterative refinement of $x$ toward compositions that more effectively meet the specified condition.

To illustrate how our multi-agent feedback-loop-based composition optimization framework operates, a simple example is provided below (**Figure 1c**). When a user provides a natural language query, for instance, "Recommend me a lead-free halide double perovskite that is thermodynamically stable", the *Condition Extractor* agent first identifies and formalizes the required constraints. Based on these extracted conditions, the *Formula Proposal Agent* generates candidate a DP composition, which is subsequently evaluated by the *Evaluator Agents*. The evaluation results are then passed to aforementioned *Text Gradient Agents,* which interprets these evaluations and suggests appropriate directions for improvement within the optimization loop. Once a refined composition is produced, the *POSCAR Matcher/Formatter* agent converts it into a structural model suitable for DFT calculations. Further details of this workflow are provided in **Section 2.1**.

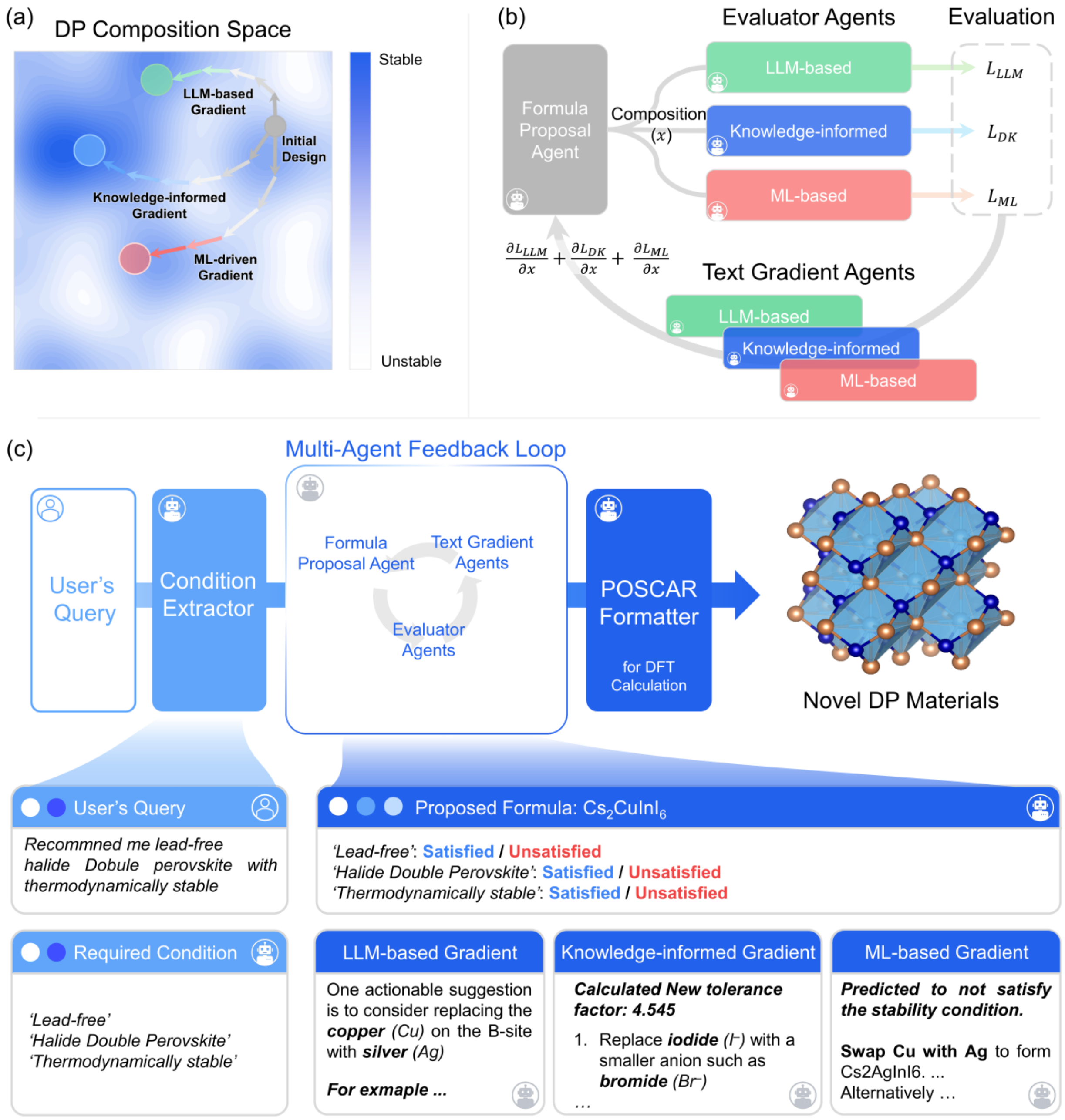

**Figure 1**. Overall architecture and illustrative example of the multi-agent LLM framework for reliable, condition-aware DP composition generation. **(a)** Conceptual illustration of how the LLM agent explores the DP compositional space by following text gradients: green for LLM-based gradients, blue for domain-knowledge-enriched gradients, and red for ML-informed gradients. **(b)**

Multi-agent feedback loop for condition-aware composition optimization. The *Formula Proposal Agent* proposes a composition $x$, *Evaluator Agents* estimate the loss based on the given condition, and *Text Gradient Agents* suggest update direction for the next iteration. **(c)** End-to-end workflow: a natural-language query is converted into a condition, the optimal DP composition is identified via the loop in (b), and the final structure is formatted and validated through downstream tools.

### 2.1 Design of Incremental Feedback Cases

Specifically, we designed three incremental cases to examine how different forms of feedback can enhance the ability of a MAS to propose DP compositions that satisfy natural language conditions. In the first case, the goal is to generate DP compositions that meet the given conditions while explicitly excluding those already reported in the Materials Project database (MP). In the second case, we further incorporate an LLM-based evaluation agent together with an LLM-derived text gradient that assesses whether the generated composition satisfies the target conditions. In the third case, we enhance the feedback loop by integrating DP-specific domain knowledge. Agents for knowledge-informed evaluation and gradient estimation are added to assess thermodynamic stability based on the new tolerance factor, a widely used empirical rule familiar to domain experts.[69] Furthermore, to investigate the potential synergy between domain knowledge and data-driven predictions, we additionally integrate an ML-based stability predictor across the cases. An overview of these three cases and their key components is provided in **Figure 2**.

For Case 1, as illustrated in **Figure 2a**, the conditions extracted from the user query are passed to the *Formula Proposal Agent*, which utilizes a pre-trained LLM to generate candidate compositions that are likely to satisfy the specified requirements. Notably, this process constitutes zero-shot generation, as the model operates without any explicit guidance regarding the physical properties or chemical feasibility of the output. To ensure novelty, a *Database Comparator* module is employed to cross-check the generated composition against existing entries in the MP database. If the proposed DP formula is found in the database, it is added to a history buffer, and the *Formula Proposal Agent* is prompted to generate a new candidate that is not included in the existing history.

This external comparator is introduced as an independent module to avoid embedding the full list of 1,494 known DPs directly into the proposal agent's prompt, which would significantly compromise the efficiency and generative performance of the LLM, as described in **Figure 2b**. When a composition is newly discovered and not found in the existing database, it is added to the list, thereby expanding the reference set beyond the original 1,494 entries. Finally, a POSCAR Matcher is employed to retrieve the most compositionally similar entry from a reference set with known structural information (e.g., POSCAR files), and the corresponding structure is adapted by substituting its constituent elements with those from the generated composition. Further details are provided in **Method** section.

In Case 2, an LLM-based evaluation agent and a text-gradient engine are introduced to evaluate and refine the composition initially generated using the same procedure as in Case 1, relying solely on the pre-trained knowledge of the LLM, as depicted by the green line in **Figure 2a**. In the first stage, the *LLM-based Evaluator* determines whether the proposed DP composition satisfies the conditions extracted from the user query. If the composition fails to meet one or more of the required conditions, it is rejected and added to the history buffer, after which the process proceeds to the gradient stage. In the second stage, the *LLM-based Text Gradient Agent* generates feedback based on the composition and the evaluation result, offering suggestions on how specific elements should be modified to better satisfy unmet conditions, as outlined in **Figure 2c**. In the subsequent iteration, the *Formula Proposal Agent* receives this text gradient, along with the history buffer, and proposes an updated composition accordingly (**Figure 2b**). This iterative process continues until all specified conditions are satisfied. Once a valid composition is obtained, the *POSCAR Matcher* is applied in the same manner as in Case 1 to prepare the structure for downstream DFT calculations.

Additionally, in Case 3, explicit domain knowledge, specifically, a new tolerance factor, is incorporated into the LLM evaluation process to assess the structural stability of the generated DP composition. The composition is first generated following the same procedure as in Case 1. It is then evaluated by two separate agents, the *LLM-based Evaluator* and the *Knowledge-Informed Evaluation*, corresponding to the green and blue evaluation paths depicted in **Figure 2a**. The former determines whether the generated composition satisfies target conditions, as in Case 2 (except for the stability constraint), and the *LLM-based Text Gradient Agent* subsequently provides feedback to improve any unmet conditions. For stability evaluation, the *Knowledge-Informed Evaluator* computes the new tolerance factor for the generated composition. The evaluator could also be automatically constructed using AutoGen,[52] which identifies the relevant equations and computational procedures from scientific literature and retrieves the necessary Shannon ionic radii from reference tables. If the stability condition is not satisfied, the *Knowledge-Informed Text Gradient Agent* then provides a corresponding knowledge-informed gradient to guide the composition toward reducing this value below 4.18, a threshold empirically associated with structural stability (**Figure 2d**).[69] The gradients from both agents are then incorporated into the *Formula Proposal Agent* in the subsequent iteration. This iterative procedure continues until all specified conditions are satisfied. Once a valid composition is obtained, the *POSCAR Formatter* updates the lattice parameters in the structural file by substituting them with values derived from the estimated ionic radii. Further details regarding the prompting strategy, the formulation of the new tolerance factor, and the POSCAR formatting procedure are provided in the **Methods** section.

To further evaluate the extensibility of the framework, we additionally integrated, for each case, a surrogate model trained on existing DP data to predict the structural stability of the generated compositions, as indicated by the red path in **Figure 2a**. The ML-based Evaluator estimates the stability of each candidate using CrabNet.[70] If the surrogate model identifies a composition as failing to meet the stability criterion, an ML-informed text gradient is generated to guide the modification of specific elements toward improved stability, following the procedure illustrated in **Figure 2e**. This experiment enables us to examine how incorporating data-driven surrogate models influences the overall generation process and complements domain-informed reasoning.

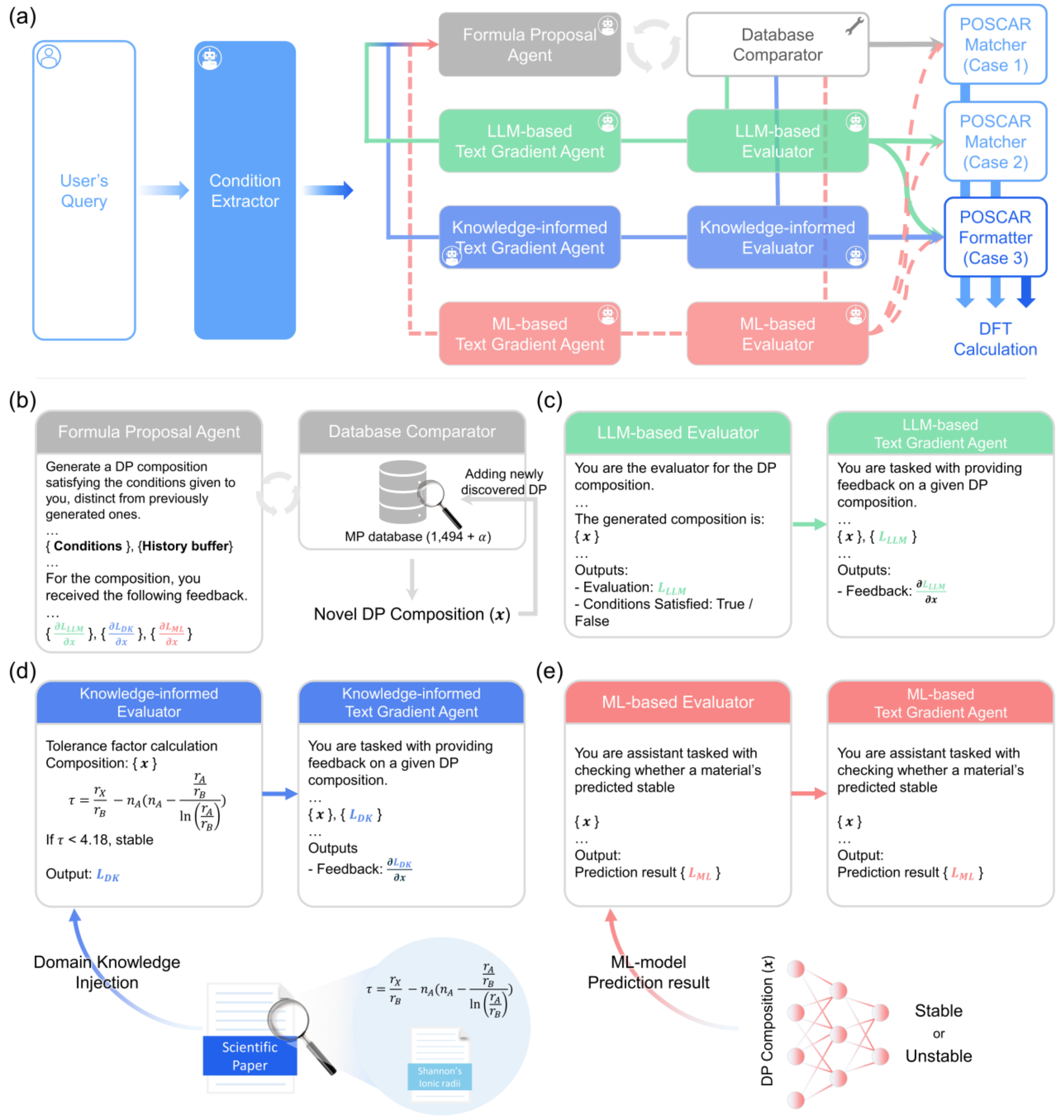

**Figure 2.** Overview of the multi-agent feedback framework and the incremental incorporation of LLM-, knowledge-, and ML-based evaluators. **(a)** Summary of the three incremental cases (Case 1–3) and the ML-based extension. Different evaluators and feedback paths, LLM-based (green), knowledge-informed (blue), and ML-based (red), are integrated stepwise across the cases. **(b)**

Interaction between the *Formula Proposal Agent* and the *Database Comparator*, which filters out compositions already present in the MP database or previously generated. **(c)** LLM-based evaluation and text-gradient feedback. The *LLM-based Evaluator* checks whether the proposed DP composition satisfies the target conditions using pre-trained knowledge, and *the LLM-based Text Gradient Agent* provides textual guidance to refine unmet conditions. **(d)** Knowledge-informed evaluation and domain-derived gradient feedback. The *Knowledge-Informed Evaluator* computes the new tolerance factor, and the *Knowledge-Informed Text Gradient Agent* produces gradients that steer the composition toward structural stability. **(e)** ML-based evaluation and ML-informed gradient feedback. The *ML-based Evaluator* calls a surrogate model to predict the structural stability of the generated composition, and the *ML-based Text Gradient Agent* provides refinement guidance when the stability criterion is not satisfied.

## 2.2. Conditions for Double Perovskites

In this study, the conditional generation performance of each LLM-based framework was evaluated using a total of five distinct queries. Each query included three types of user-specified conditions: (1) property conditions, (2) element conditions, and (3) perovskite-type conditions. The details of each query can be found in **Table 1**. The property conditions, describing the desired properties of the generated materials, were applied consistently across all five queries.

The property condition was defined as thermodynamic stability, which reflects energetic favorability and equilibrium stability of a material.[71,72] It is quantitatively evaluated using the formation energy and energy above hull. DPs with negative formation energy and zero energy above hull were classified as **stable**, while those with energy above hull below 0.05 eV/atom were regarded as **metastable**, reflecting their possible existence under experimentally accessible conditions.[73]

The element condition, applied in queries 4 and 5, controls the inclusion or exclusion of specific elements in proposed DPs. Such compositional constraints are crucial for tailoring material functionality—for instance, incorporating rare-earth elements can enhance optical performance,[74,75] while excluding lead mitigates environmental and health concerns.[76,77] This criterion was therefore introduced to enable targeted material discovery. And, all queries also included a perovskite-type condition, which specifies the intended type of perovskite—halide, chalcogenide, or oxide—based on the X-site anion of the proposed composition. This condition serves to guide the *Formula Proposal Agent* toward generating materials that belong to the desired perovskite family. For each query, 30 DPs were generated (150 per case), which were subsequently used to evaluate the performance of each case.

**Table 1.** Five queries used in this study. Each query contains (1) property conditions related to thermodynamic stability; (2) elements conditions related to whether specific elements are contained in double perovskite; and (3) perovskite-type conditions specifying the perovskite type, such as halide double perovskite, chalcogenide double perovskite, and oxide double perovskite.

| **Index** | **Queries** |
|---|---|
| Query 1 | Recommend me thermodynamically stable[1] halide double perovskites[3] |
| Query 2 | Recommend me thermodynamically stable[1] chalcogenide double perovskites[3] |
| Query 3 | Recommend me thermodynamically stable[1] oxide double perovskites[3] |
| Query 4 | Recommend me thermodynamically stable[1], lead-free[2] halide double perovskite[3] |
| Query 5 | Recommend me thermodynamically stable[1] oxide double perovskite[3] that contains rare earth elements[2] |

## 2.3 Analysis for Generated DPs

In this section, we evaluate the DPs generated in each case, focusing on (i) property-condition satisfaction and (ii) elemental and perovskite-type condition satisfaction. Case 1 corresponds to zero-shot LLM generation without any feedback; Case 2 incorporates LLM-based evaluation and text-gradient feedback; and Case 3 further integrates knowledge-informed evaluation and domain-derived gradients based on the new tolerance factor. By comparing these cases, we aim to disentangle how different forms of feedback, LLM-based reasoning, domain-informed evaluation, and their interaction, shape the reliability, constraint adherence, and physical plausibility of LLM-driven composition generation.

**Figure 3** summarizes the degree to which the generated DPs satisfy the property-related condition. Generated compositions for all cases exhibit negative formation energies (**Figure 3a**), demonstrating that the LLM is capable of proposing thermodynamically favorable material candidates. Furthermore, beyond compositional feasibility, we evaluated the phase stability relative to competing phases of each DP by computing its energy above hull. Following standard conventions, DPs with energy above hull < 0.05 eV/atom were classified as metastable, and those with 0.00 eV/atom were considered stable state. (**Figure 3b-c**). Among all cases, Case 3 produced the most stable candidates, attaining an average value of approximately 0.09 eV/atom, about 30% lower than those obtained in Cases 1 and 2. Correspondingly, the fraction of stable and metastable DPs increased steadily from Case 1 (~43%) to Case 2 (~48%) and Case 3 (~54%), highlighting the beneficial impact of LLM-based feedback and domain-informed evaluation on improving property-condition satisfaction during DP generation. In contrast, earlier GAN-based approaches addressing the similar task of generating stable DPs reported ~27% stability-compliant candidates, while our LLM-based framework achieves markedly higher fractions, showing that the chemically grounded token-level reasoning of LLMs provides a stronger inductive bias for generating syntactically and structurally valid materials candidates.[21] Detailed property distributions for each individual query are provided in **Figure S1-S3** in **Supplementary Note 1**.

The improvement observed from Case 1 to Case 2 can be attributed to the introduction of LLM-driven self-evaluation rather than any difference in the underlying knowledge base. Both cases rely exclusively on the pretrained knowledge of the LLM; however, Case 1 performs a single-pass, zero-shot generation without any evaluative mechanism, whereas Case 2 activates an additional feedback loop through the LLM-based evaluation and the text-gradient engine. This self-evaluation capability enables the model to assess whether the generated composition satisfies the target property condition and to iteratively refine the proposal through natural-language gradient feedback. These observations align with recent reports showing that LLMs exhibit emergent abilities for self-diagnosis and self-correction, even without task-specific training.[62,78]

Additionally, Case 3 further strengthens this mechanism by incorporating DP-specific domain knowledge into the textual feedback loop. The knowledge-informed text gradient provides a more effective search direction than the purely LLM-based gradient, particularly for guiding the model toward thermodynamically favorable regions of the DP compositional space. Interestingly, although the new tolerance factor used as the domain-knowledge criterion in this study is not directly correlated with energy above hull across diverse DPs,[79] its integration nevertheless biases the LLM's refinement process toward compositions with lower energy above hull values. This suggests that the combination of domain heuristics with the LLM's learned priors produces synergistic effects, enabling the model to navigate the stability landscape more reliably than with LLM-based feedback alone.

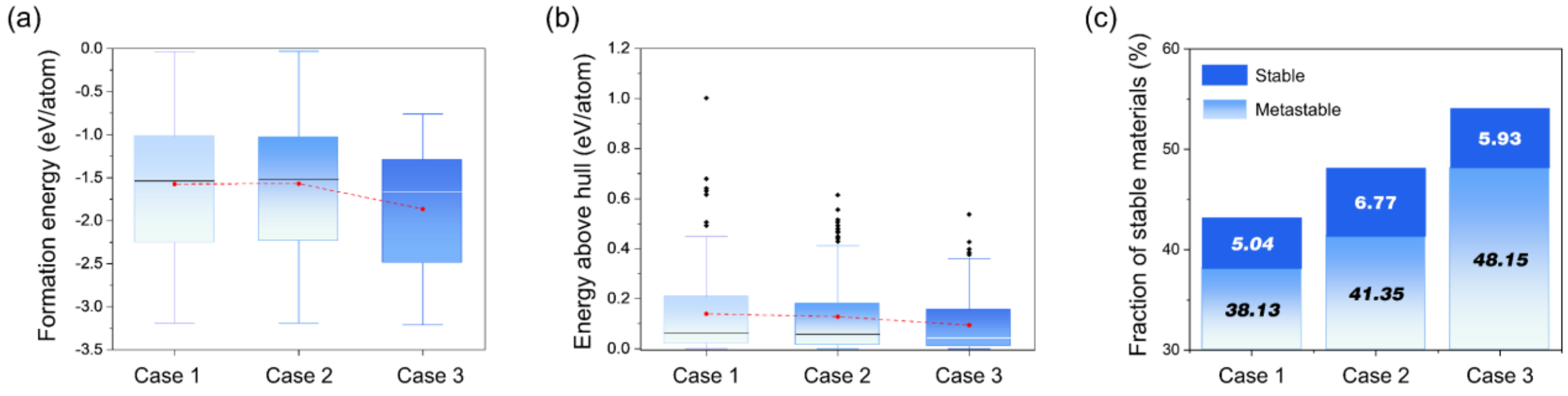

**Figure 3.** Results of thermodynamic conditions for generated DPs in Cases 1-3. **(a)** Distribution of formation energies for all DPs generated in each case. **(b)** Distribution of energy above hull for all DPs generated in each case. **(c)** Fraction of generated DPs classified as stable or metastable states in each case.

For the perovskite-type condition, the iterative feedback driven by the knowledge-informed gradient in Case 3 can sometimes become overly restrictive. As the *Formula Proposal Agent* repeatedly adjusts the composition to satisfy the tolerance-factor constraint, the model occasionally generates candidates that violate the intended perovskite-type condition. An example of such a violation scenario is shown in **Supplementary Note 2**. Specifically, in Cases 1 and 2, all generated DPs satisfied the intended element and perovskite-type conditions across all queries. For Case 3, approximately 98.5% of the generated DPs met these predefined perovskite-type and elements conditions as shown in **Figure 4a**. As rejections accumulate and the history buffer expands, the feasible compositional search space progressively narrows, increasing the likelihood that the *Formula Proposal Agent* generates unintended compositions. This tendency is particularly evident in Case 3, where the explicit new tolerance factor criterion leads to more frequent rejections, on average about 6.1 per generated DP compared with 0.6 in Case 2 as presented in **Figure 4b**.

This effect was most pronounced under the chalcogenide DP condition (query 2), where violation of the perovskite-type condition was observed, as shown in **Figure 4c**. According to the new tolerance factor criterion, compositions with larger X-site ionic radii are less likely to satisfy the stability threshold (see **Methods** section). The X-site anions of chalcogenide DPs, $Se^{2-}$ (1.98 Å) and $S^{2-}$ (1.84 Å), are relatively larger than those of halogen or oxygen ions,[80] which caused the *Knowledge-Informed Evaluator* in Case 3 to reject the proposed compositions more frequently. Consequently, Case 3 recorded the highest number of rejections, averaging approximately 27.7 per generated DP in query 2. This excessive rejection process eventually led to violations from the intended perovskite-type condition. Detailed rejection counts per query are provided in **Supplementary Note 3**.

Similarly, under the halide DP conditions (queries 1 and 4), the generated DPs in Case 3 showed an elemental trend favoring X-site anions with smaller ionic radii, $F^-$ (1.33 Å) < $Cl^-$ (1.81 Å) < $Br^-$ (1.96 Å) < $I^-$ (2.20 Å). In contrast, Cases 1 and 2 rarely selected fluorine as the X-site element when generating halide DPs. This tendency likely stems from the LLM's pre-training process, which was verified using knowledge-probing techniques. Details of this analysis are provided in the **Supplementary Note 4**.

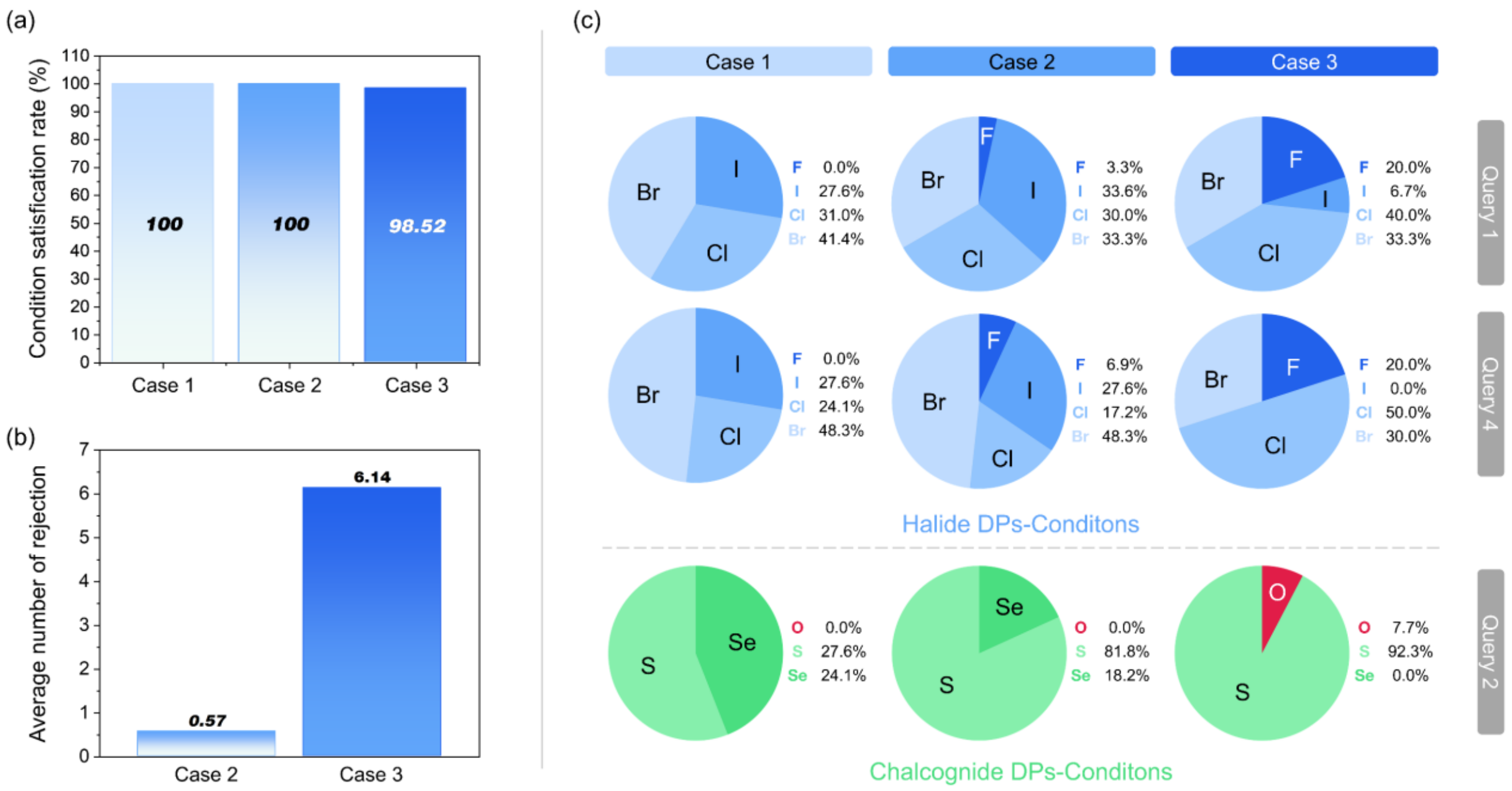

**Figure 4. (a)** Satisfaction rates of elemental and perovskite-type conditions for each case. **(b)** Average number of rejections per finally accepted composition for cases 2 and 3. **(c)** Pie charts showing the distribution of X-site ions under halide and chalcogenide perovskite-type conditions for each case.

To further understand how generated DPs evolved during the multi-agent feedback process, we investigated the optimization process of proposed formulas in Case 2 and Case 3. This analysis examines how initially proposed compositions were iteratively refined through agent interactions, focusing on incremental elemental substitutions and their influence on thermodynamic stability.

As shown in **Figures 5a-c**, Case 2 primarily optimized compositions through substitutions at the B- and B′-sites, while the A-site element generally remained unchanged. For instance, in query 1, the initially proposed $Rb_2InCuI_6$ evolved through substitutions at the B/B′-sites to yield $Rb_2AgSbF_6$. A similar trend was observed for queries 2 and 4. Most recommended DPs in Case 2, such as Cs- or Rb-based halides and Ba-based chalcogenides, were confined to well-studied compositional families, indicating that gradient-based feedback encouraged local optimization within familiar chemical domains.

In contrast, Case 3, which incorporated new tolerance factor-based feedback, explored a broader range of elemental combinations across all sites. Compared with Case 2, which proposed Cs- or Rb-based DPs, Case 3 recommended new elemental combinations, incorporating lighter alkali-metal ions (Na, K, Li) at the A-site and additional B/B′-site cations such as Tl, Cd, Pb, Ba and Sr to satisfy stability feedback. In short, the knowledge-informed gradient broadened the search landscape within the DP composition space, enabling exploration beyond the local B/B'-site adjustments observed in Case 2 and guiding the agent toward more thermodynamically favorable optima. Distributions of A-site, B-site, B'-site, and X-site elements per query for Cases 1-3 are provided in **Figures S6-S8** in **Supplementary Note 5**.

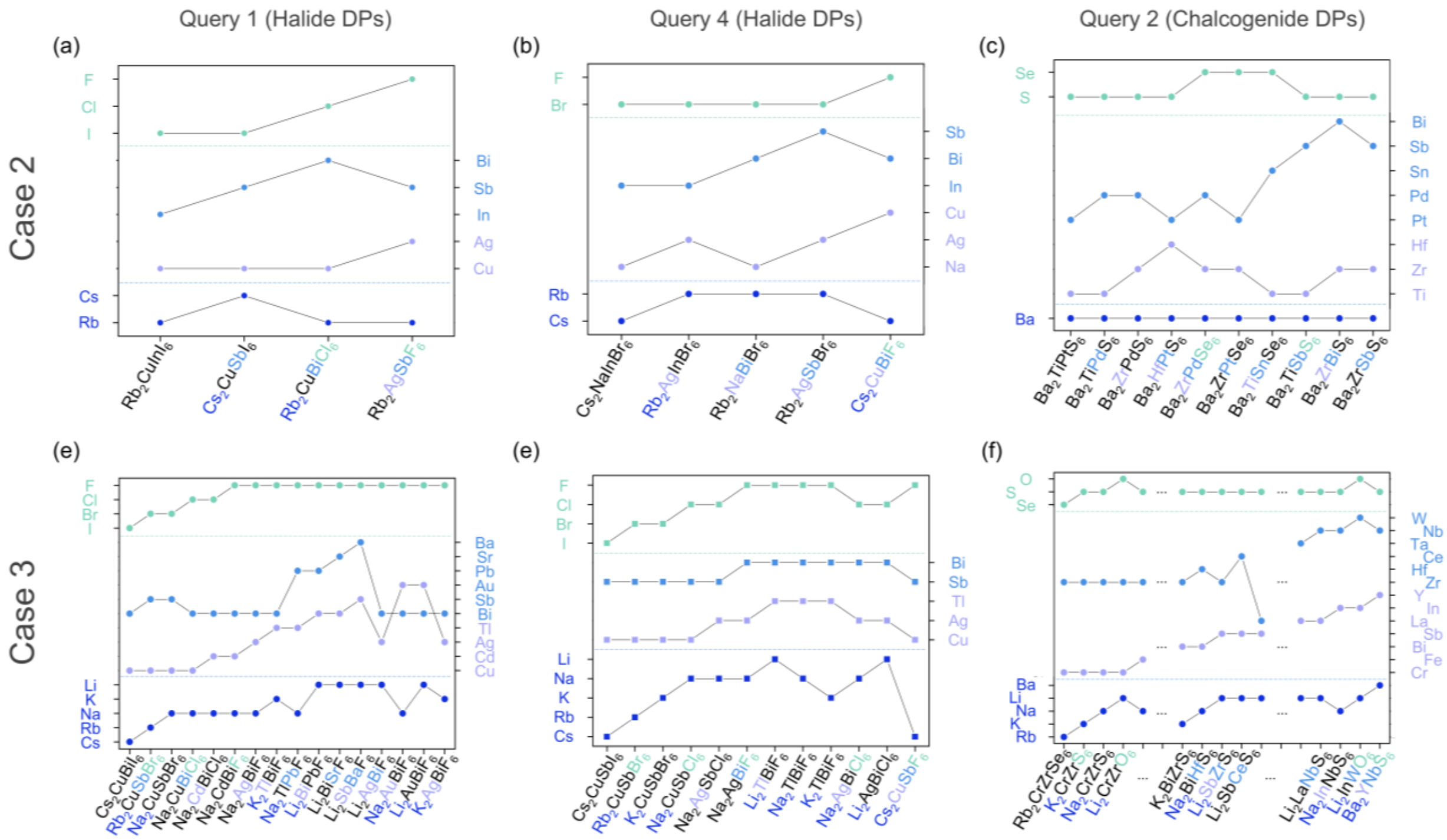

**Figure 5.** Evolution of compositions in representative examples from Case 2 and Case 3. The shown cases correspond to formulas that experienced the highest numbers of rejections in each query, illustrating how iterative feedback refined the elemental combinations throughout the optimization process. For each panel, scatter plots depict compositional shifts between the proposed and generated DP proposals at the A-site cations (dark blue), B/B′-site cations (light purple and blue), and X-site anions (light green). **(a)–(c)** Representative results for Case 2 (queries 1, 4, and 2. **(d)–(f)** Representative results for Case 3 (queries 1, 4, and 2). Oxide DPs were excluded due to their relatively low rejection counts.

As the elemental combinations evolved through iterative feedback, the satisfaction of the property condition, defined by formation energy and energy above hull, also varied accordingly. As shown in **Figure 6**, iterative feedback generally lowered the formation energies of the generated DPs compared with both the initially proposed DPs and the average formation energies of all intermediate proposals. For example, in Case 2 (query 1), the formation energy decreased from –0.77 eV/atom for the initial $Rb_2CuInI_6$ to –2.55 eV/atom for the final $Rb_2AgSbF_6$. This tendency was consistently observed in both Case 2 and Case 3, confirming that feedback effectively guided the search toward more energetically favorable configurations.

However, the behavior of energy above hull values was more complex. Among the generated DPs, some showed clear improvement over the initial proposals, for instance, $Ba_2YNbS_6$ in Case 3 (query 2) exhibited a reduced energy above hull of 0.084 eV/atom compared with 0.23 eV/atom for $Rb_2ZrCrSe_6$, while others were lower than the average of intermediate candidates but still higher than the initial compositions, such as $Cs_2CuSbF_6$ (0.155 eV/atom) compared with 0.061 eV/atom for $Cs_2CuSbI_6$ (query 4). This inconsistency arises from the fundamental difference between the two thermodynamic metrics. Formation energy evaluates a material relative to its constituent elements, whereas the energy above hull measures its stability against all possible competing phases within the same compositional space. Consequently, achieving a lower energy above hull requires exploring a much broader thermodynamic landscape that includes numerous alternative configurations.[81]

Overall, we found that, perhaps surprisingly, the formation energy decreased in all cases, demonstrating that the proposed workflow consistently optimizes DP composition through iterative feedback. In contrast, energy above hull optimization proved more challenging. While several compositions achieved lower hull values than their initial proposals, others did not improve monotonically. This reflects the inherent difficulty of reducing the hull energy, which depends on stability against all competing phases in the compositional space, compared with the simpler element-based stability captured by the formation energy. Even so, although not all compositions improved, the overall distribution of hull energies shifted downward in Cases 2 and 3, indicating that the incorporation of domain knowledge provides a meaningful advantage for stability-oriented exploration, as shown in **Figure 3**.

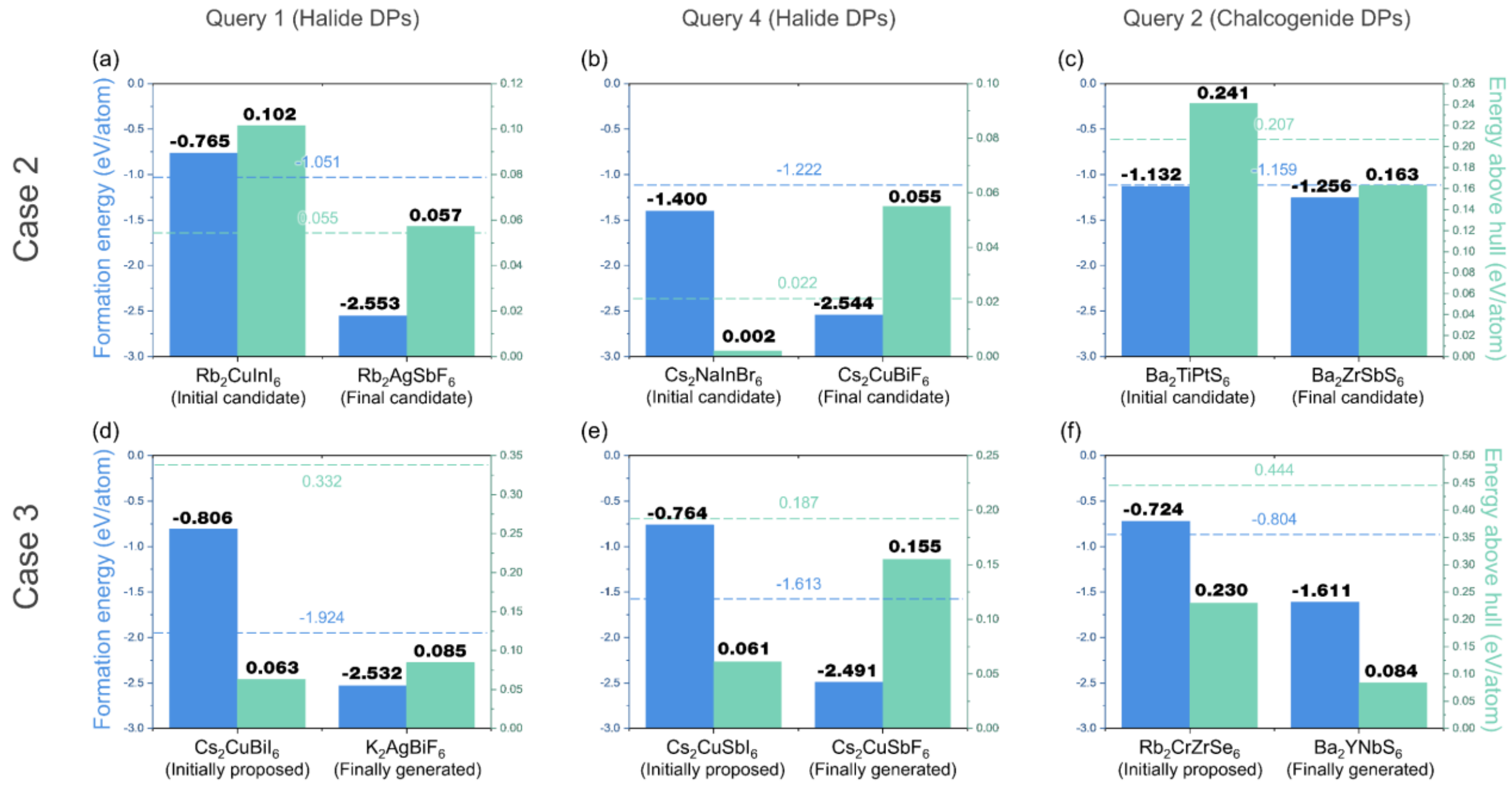

**Figure 6.** Evolution of formation energy and energy above hull in representative examples from Case 2 and Case 3. The shown cases correspond to formulas that experienced the highest numbers of rejections in each query, demonstrating how iterative feedback influenced the thermodynamic stability. Each bar plot compares the thermodynamic properties of the initially proposed DPs (initial candidate), the final generated DP (final candidate), and the average of intermediate but rejected DPs. Blue bars denote formation energies, light-green bars represent energies above the hull, and dashed lines indicate mean values of rejected DPs.

Collectively, Case 3, enabled by the knowledge-informed text gradient, generated a larger fraction of materials that satisfied the property condition compared with Cases 1 and 2. This indicates that the domain-informed feedback facilitated more effective exploration of the DP composition space, guiding the model toward thermodynamically more favorable optima on average. However, this mechanism was not without limitations. The stricter feedback introduced by the *Knowledge-Informed Text Gradient Agent* also led to a higher rejection rate. As the history buffer expanded through repeated rejections, the *Formula Proposal Agent* more frequently produced redundant or unintended compositions, making it more challenging to satisfy the elemental and perovskite-type constraints for certain chalcogenide DPs. Such behavior is consistent with well-documented challenges in physics-informed and knowledge-guided machine learning, where simultaneously optimizing multiple constraints or loss terms often leads to gradient conflicts and trade-offs among competing objectives.[82] These issues are therefore not unique to our framework, and future work may leverage advances in multi-objective optimization and constraint-balancing strategies to further mitigate such interactions. Further discussion of these interactions and their implications is provided in **Section 2.5**.

### 2.4 Integration of Surrogate Models

In this section, we examine the influence of surrogate model-based approaches in generative materials discovery. Because surrogate models typically perform well within their ID regimes but degrade substantially in OOD regions, we first classified all generated DPs into ID and OOD spaces. Based on this separation, we then compared the performance of the ML-integrated configurations (**Case 1 + ML**, **Case 2 + ML**, and **Case 3 + ML**) against their MAS frameworks (**Case 1**, **Case 2**, and **Case 3**). A schematic overview of these configurations is provided in **Figure 2a**.

We used a CrabNet-based surrogate model as a stability classifier. CrabNet was trained on DP entries from the Materials Project, where each composition was labeled as **'stable' or 'metastable' (**formation energy < 0 eV/atom and energy above hull < 0.05 eV/atom) or '**unstable**' otherwise. The model takes the raw chemical formula as input and embeds it into a composition-aware attention framework, enabling the capture of element–element interactions without requiring structural descriptors.

To separate ID from OOD regions with respect to the CrabNet training data, we computed the Elemental Mover's Distance (ElMD)[83] between each generated DP and the compositions present in the original CrabNet dataset. A smaller ElMD value therefore reflects higher chemical similarity. We quantified each composition's proximity to the training distribution using the average of its ten nearest neighbors (Top-10 ElMD). Compositions with low Top-10 ElMD values were classified as ID, whereas those with high values were designated as OOD.

**Figure 7** illustrates a clear distinction between the ID and OOD regions. DPs generated in queries 1, 3, and 4 exhibited small Top-10 ElMD values (approximately 0.3-0.6), indicating strong compositional similarity to the training data. In contrast, queries 2 and 5 showed much larger ElMD values (around 1.6), confirming that these compositions lie well outside the model's training chemical space. Consistent with ElMD trends, the surrogate model maintained high predictive performance for ID cases (weighted F1 scores of approximately 0.85-0.88) but dropped markedly for OOD cases (around 0.4-0.5). This discrepancy highlights the limited extrapolation capability of the surrogate model beyond its training distribution. Detailed performance metrics are summarized in **Supplementary Note 6.**

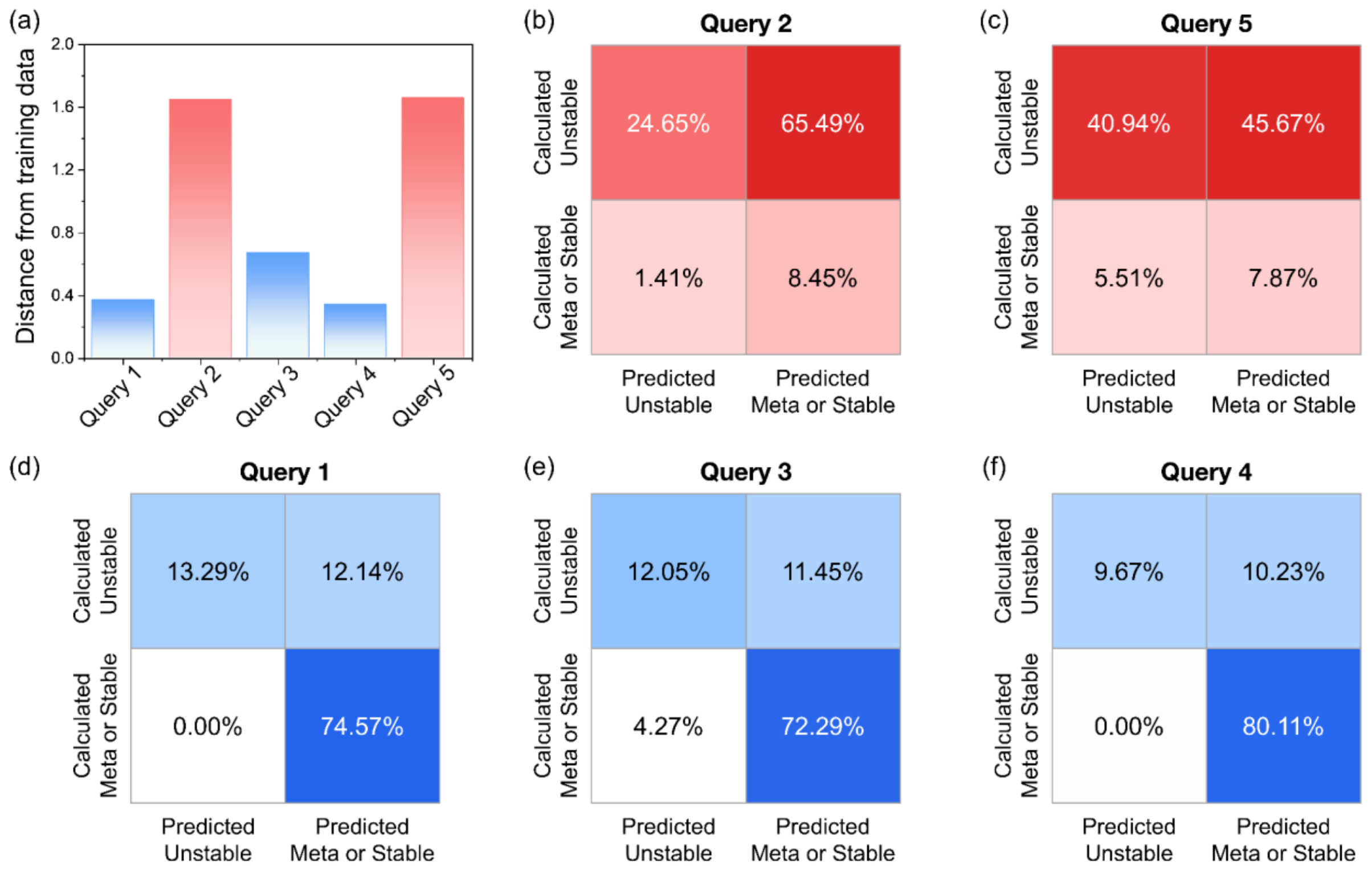

**Figure 7.** Classification performance of the surrogate model. **(a)** Average of the 10 closest elemental mover distances between training data and generated double perovskite compositions, for queries 1–5. **(b)–(f)** Confusion matrices for query 2, query 5, query 1, query 3, and query 4, respectively.

The consequences of this OOD limitation become clear once the surrogate model is incorporated into the generative frameworks. In the ML-integrated settings (Case 1 + ML, Case 2 + ML, and Case 3 + ML), the thermodynamic stability of each generated DP was re-evaluated to determine how ML-based text gradients influence property consistency.

From the perspective of formation energy, **Figure 8a** shows that only one composition across all generated DPs—$Rb_2CaBiSe_6$ (0.147 eV/atom), obtained in Case 2 + ML under Query 2, exhibited a positive formation energy. Notably, this case corresponds to an OOD query. This example highlights that, unlike the MAS frameworks, the ML-integrated framework can occasionally violate the property condition due to inaccurate surrogate predictions in OOD regimes.

The reduced accuracy of the surrogate model in OOD regions was also evident in the energy above hull distributions. As illustrated in **Figures 8b-c**, DPs generated under OOD queries (e.g., queries 2 and 5) using ML-integrated frameworks exhibited higher energy above hull values than those obtained from LLM-based frameworks. For instance, in query 2, Case 2 + ML yielded an average energy above hull of 0.38 eV/atom, nearly twice that of Case 3 without ML (0.19 eV/atom). This degradation in thermodynamic stability underscores the surrogate model's limited reliability when operating beyond its training distribution. In contrast, for ID queries, the difference between ML-integrated and LLM-only frameworks was marginal, suggesting that the adverse effects of surrogate integration become pronounced primarily under OOD conditions.

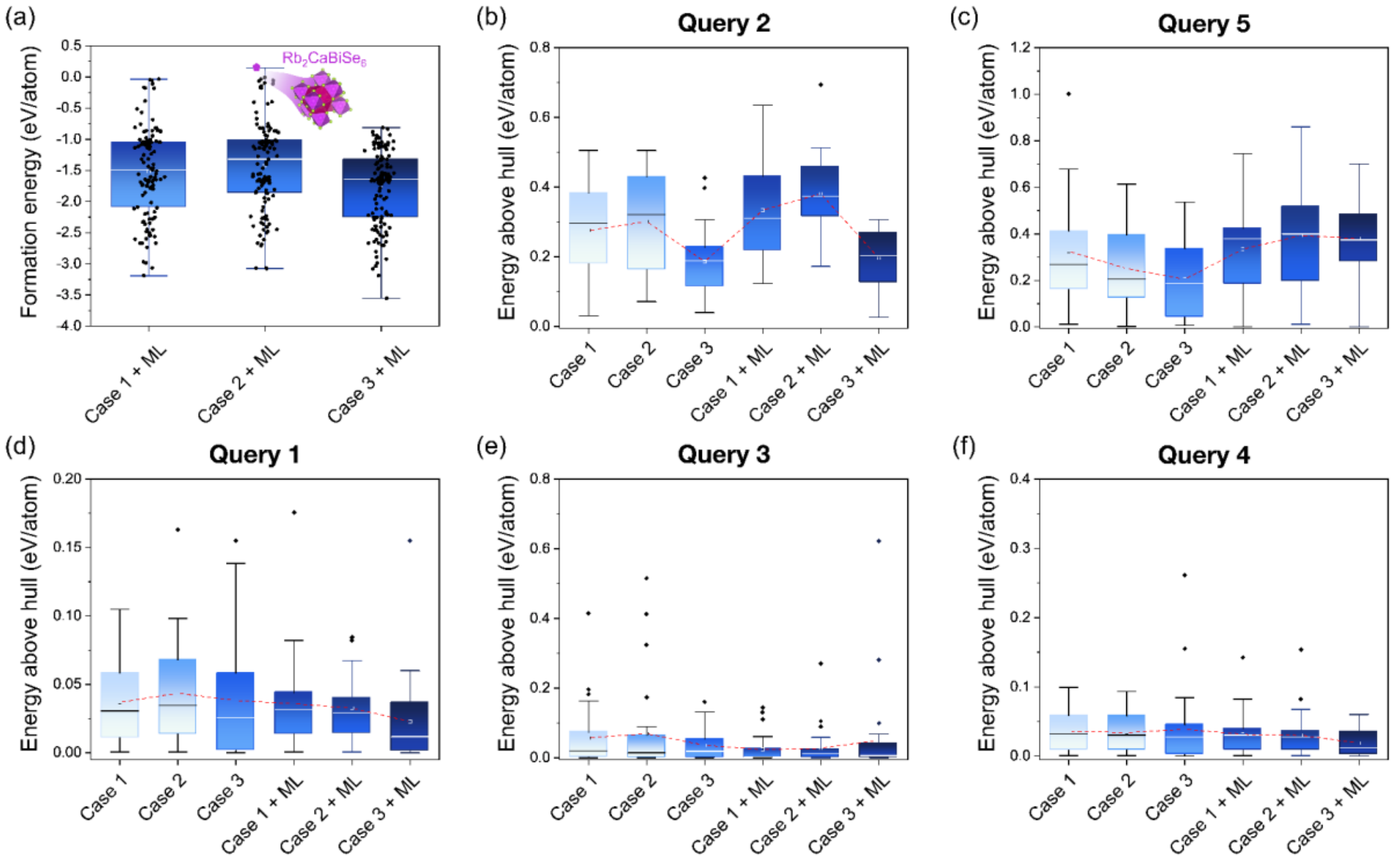

**Figure 8**. Distribution of thermodynamic properties. **(a)** Distribution of formation energies for double perovskites generated in Case 1 + ML, Case 2 + ML, and Case 3 + ML. **(b)-(f)** Distributions of energy above hull values for double perovskites generated in all cases for queries 2, 5, 1, 3, and 4, respectively. Only double perovskites that satisfy the elemental and perovskite type conditions were considered for rigorous analysis.

These results highlight that, while the ML-based gradient can effectively guide the search toward favorable regions of the DP composition space when the generated candidates lie within the ID domain of the surrogate model, it may exert a misleading or even detrimental influence when operating in OOD regions, where the model's predictions are inherently unreliable, even with additional training data. The issue becomes especially critical in fields where high-quality material data are scarce or costly to obtain. Therefore, careful consideration of the intended design space becomes critical. To maintain reliable guidance from the surrogate model, the design targets should be selected such that the generated compositions remain within a near-ID region where predictive accuracy can be sustained. Consequently, a well-constructed surrogate model with an appropriately defined training domain is essential for enabling effective generative exploration driven by the ML-based gradient.

### 2.5 Limitations and Future Work

As demonstrated in the preceding analyses, our multi-agent framework provides several important insights into the role of different feedback modalities in guiding DP composition generation. First, incorporating knowledge-informed feedback through the new tolerance factor enabled the system to explore the DP composition space more effectively, producing compositions that were, on average, thermodynamically more favorable than those obtained through LLM self-reasoning alone. Second, integrating ML-based feedback improved stability-oriented optimization within the ID regions of the surrogate model but simultaneously revealed clear limitations in OOD regimes, where reliable predictions occasionally misdirected both the promise and the constraints of LLM-based, knowledge-informed, and ML-based gradients. Building on these results, the following outlines the key limitations identified for each gradient type and discusses potential strategies to address them in future work.

First, we found that the strict constraint imposed by the knowledge-informed gradient led to a higher rejection rate and, in some cases, caused the framework to generate compositions that violated other intended conditions. This indicates that the strong emphasis on stability diminished the influence of gradients associated with other constraints, such as the perovskite-type requirement. These observations underscore the need for a more principled approach to balancing multiple gradients so that their synergistic benefits can be fully realized. In this work, we simply concatenated the gradients and merged them into the prompt for the *Formula Proposal Agent*; however, more sophisticated integration strategies are likely to yield improved performance. One promising direction is the introduction of an administrative agent that aggregates and reconciles feedback from multiple evaluators. Such a mechanism would enable the framework to more effectively harmonize the LLM's pretrained priors with domain- and physics-based constraints, thereby enhancing the consistency and reliability of the generated compositions.

Second, the proposed multi-agent framework consistently improved the formation energy, however, the energy above hull value was not consistently improved. The energy above hull is only indirectly guided and does not explicitly account for phase competition in the composition space. To overcome this limitation, one promising direction would be to incorporate an additional gradient engine agent that queries open materials databases, evaluates the thermodynamic stability of proposed formulas against competing phases, and feeds this information back to steer generation toward candidates to lower energy above hull.[48,59] Such database-grounded evaluation could improve the identification of globally stable materials and yield more thermodynamically consistent optimization.

Third, the ML-based text gradient was largely ineffective in OOD regions of the DP composition space. Addressing this limitation will require improving the robustness of surrogate models beyond their training distribution. While building a universal surrogate model for the entire DP domain remains challenging due to fundamental extrapolation limits, an alternative direction is to explore mechanisms that adaptively enhance the surrogate model's reliability in regions relevant to a given design task. More broadly, approaches that combine selective data acquisition with iterative model refinement may provide a promising path for extending the usefulness of ML-based gradients, particularly when aiming for novel materials discovery.

Beyond the present focus on DPs, the proposed framework can be naturally extended to a broader range of materials systems. Future work may explore its application to single perovskites, anti-perovskites, or other structurally diverse chemistries, thereby assessing the generality of the multi-agent approach across different crystal families. Likewise, although this study centered on thermodynamic stability as the primary design objective, the same gradient-driven workflow may be adapted to optimize electronic, optical, or mechanical properties, provided that appropriate evaluators or domain rules are available. Importantly, such extensions will require not only expanding the framework to new materials and properties but also carefully determining which forms of domain knowledge and which types of surrogate models should be incorporated as feedback sources, as their effectiveness depends strongly on the characteristics of the targeted design space and the nature of the underlying optimization objective. Thoughtful selection and integration of these information sources will therefore be essential for ensuring reliable performance as the framework is expanded to more complex materials and property landscapes.

## 3. Conclusion

In this work, we introduced a multi-agent, text gradient-driven framework for conditional generation of DP materials, integrating three complementary forms of feedback: LLM-based evaluative feedback, DP-specific domain-knowledge–informed feedback, and ML prediction-based feedback. To assess their effectiveness, firstly, we systematically compared three configurations- pure generation without feedback (Case 1), generation with LLM-based feedback (Case 2), and generation with domain-knowledge-informed feedback (Case 3). The results show that iterative feedback and domain-knowledge integration substantially improve satisfaction of property-related conditions without requiring additional training data. Overall, the proposed framework achieved over 98% satisfaction of elemental and perovskite-type constraints and generated up to ~54% stable or metastable DPs, outperforming both the LLM-only baseline (~43%) and a previously reported GAN-based generative baseline (~27%).

Additionally, our analysis revealed that ML-based text gradients provide meaningful guidance within the ID regions of the surrogate model, but their benefits cannot be guaranteed in OOD regimes. This trend was evident in the stability-oriented generation results: for ID-aligned chemistries such as common halide and oxide DPs, the ML-based gradient substantially improved performance, leading to lower average energy above hull values. In contrast, for OOD cases, such as oxides containing rare-earth elements, which lie far outside the surrogate's training distribution, the same ML-based feedback degraded performance, yielding higher average hull energies. These findings highlight that while ML-based gradients can significantly enhance conditional generation within well-represented regions of the composition space, their reliability sharply deteriorates in OOD regimes, indicating that the synergistic interaction between domain knowledge and ML-based predictors cannot be guaranteed when the surrogate operates outside its distributional support.

Based on these analyses, several directions remain for further improving the framework. First, developing principled strategies for balancing multiple gradients, determining the appropriate influence each evaluator should exert, will be essential, analogous to loss-weighting strategies in multi-objective optimization or physics-informed machine learning. Second, for the ML-based gradient, iterative refinement strategies that expand the surrogate model's distributional coverage, handling OOD issue, offer a more practical path toward enabling ML-informed gradients to contribute readily to the discovery of novel materials.

This study provides the first systematic analysis of DP generation using a multi-agent LLM framework incorporating proposed concepts of knowledge-guided and ML-based text gradient, and elucidates how different types of feedback influence the generation trajectory. The proposed approach can be extended beyond DPs to single perovskites, and anti-site-ordered phases, and other material families, as wells as to objectives beyond thermodynamic stability. Such extensions, however, will require careful decisions regarding which forms of domain knowledge, which surrogate models, and which evaluators are most appropriate for a given design problem. By offering both empirical insights and methodological foundations, we expect this framework to serve as a versatile blue-print for MAS-guided generative design across a wide range of materials systems for advancing sustainable technologies.

## 4. Methods

### New Tolerance Factor Calculation

For evaluating the structural stability of the suggested DP composition, the new tolerance factor is applied, which is calculated following the equation below:

$$\tau = \frac{\mathrm{r_X}}{r_B} - n_A(n_A - \frac{r_A/r_B}{\ln(r_A/r_B)})$$

The new tolerance factor calculation requires oxidation state and ionic radius of each element composing the DP. For determining the radius, the process actively matches charge neutrality based on oxidation state and coordination number and finds Shannon ion radius on the reference table.

### POSCAR Matcher

For the DFT calculation, a POSCAR input file must be constructed, containing information about the lattice vectors, atomic species and their counts, and the fractional atomic coordinates within the unit cell. In Case 1 and Case 2, only the composition is generated by the model; thus, no structural information is directly available. To address this, we identified the most compositionally similar structure from a reference list of DP compositions, for which corresponding POSCAR files are already known. The reference dataset of 1,494 known DP compositions was obtained from the MP.

The detailed procedure is outlined as follows. First, we embedded all reference compositions into a high-dimensional vector space using the OpenAI Embedding model, such that each composition was represented as a semantic vector. The generated composition from the *Formula Proposal Agent* was then embedded using the same model, and its ten nearest neighbors in the embedding space were retrieved using cosine similarity. These ten candidate compositions were provided as input to the *Reference Recommender*, an LLM-based chain (powered by GPT-4o), which selected the most structurally appropriate composition using its pre-trained domain knowledge. Once the most similar reference composition was selected, its corresponding POSCAR file was retrieved, and the atomic species were substituted with those of the generated composition, while all other structural parameters were preserved.

This retrieval-augmented approach was adopted to overcome the input length limitation of LLMs; directly including all reference compositions in a single prompt would significantly degrade the model's inference efficiency and accuracy.

**POSCAR Formatter**

For Case 3, information about the elements such as oxidation state, coordination number, and Shannon ionic radius is available, which allows the calculation of the new tolerance factor as shown below.

$$\tau = \frac{r_X}{r_B} - n_A\left(n_A - \frac{r_A/r_B}{ln(r_A/r_B)}\right)$$

Therefore, we have substituted standard POSCAR format for cubic DP structure with the calculated lattice parameter a as below.

$$a = 2 \times (r_B + r_X) \times \sqrt{t}$$

This approach enables the generation of candidate structural inputs appropriate for subsequent DFT calculations, without relying on additional LLMs.

**Construction of Multi-agent LLM Workflow and Prompting for each agent**

In this research, the effectiveness of different types of feedback (gradients) within a multi-agent LLM workflow was examined to optimize DP compositions. Feedback was derived from three sources: the LLM itself, expert domain knowledge, and integrated ML tools. For the implementation of this research, structured prompting was employed to ensure a clear division of labor across agents, with roles and inputs/outputs explicitly specified. The workflow was implemented in Python 3.11 using LangChain and LangGraph, and all LLM calls relied on the o3-mini reasoning model.

**DFT Calculation**

Spin-polarized first-principles calculations were performed using the Vienna Ab initio Simulation Package (VASP) with the projector-augmented-wave (PAW) method. Exchange–correlation effects were treated within the Perdew–Burke–Ernzerhof (PBE) formulation of the generalized gradient approximation (GGA). A plane-wave energy cutoff of 520 eV was applied. Electronic self-consistency was achieved using a convergence criterion of 1 × 10-6 eV, and ionic relaxations were carried out with the conjugate-gradient algorithm until the maximum force on any atom was below 0.05 eV Å-1. A Hubbard U correction was applied to the localized d orbitals. All input parameters and calculation settings were generated automatically via the MPRelaxSet class in Python Materials Genomics library.

# Supporting Information

## Enhanced Conditional Generation of Double Perovskite by Knowledge-Guided Language Model Feedback

Inhyo Lee[1,†], Junhyeong Lee[1,†], Jongwon Park[1], KyungTae Lim[2,*], and Seunghwa Ryu[1,3*]

**Affiliations**

[1] Department of Mechanical Engineering, Korea Advanced Institute of Science and Technology (KAIST), 291 Daehak-ro, Yuseong-gu, Daejeon 34141, Republic of Korea

[2] Graduate School of Culture Technology, Korea Advanced Institute of Science and Technology (KAIST), 291 Daehak-ro, Yuseong-gu, Daejeon 34141, Republic of Korea

[3] KAIST InnoCORE PRISM-AI Center, Korea Advanced Institute of Science and Technology (KAIST), Daejeon, 34141, Republic of Korea

Corresponding author: ryush@kaist.ac.kr, ktlim@kaist.ac.kr

## Supplementary Note 1. Property distributions for each individual query

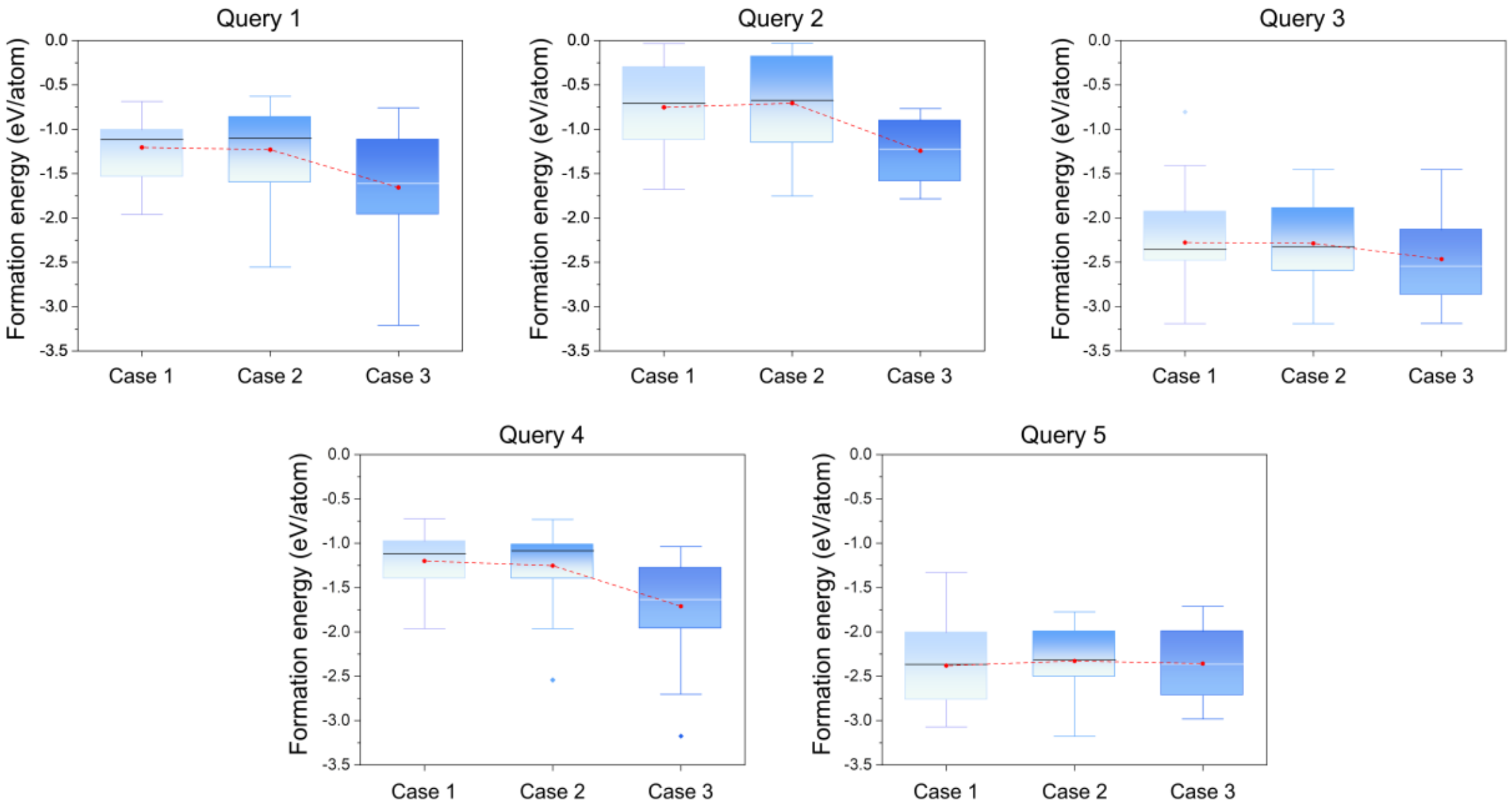

**Figure S1.** Results of formation energy for generated DPs in cases 1-3 across all queries.

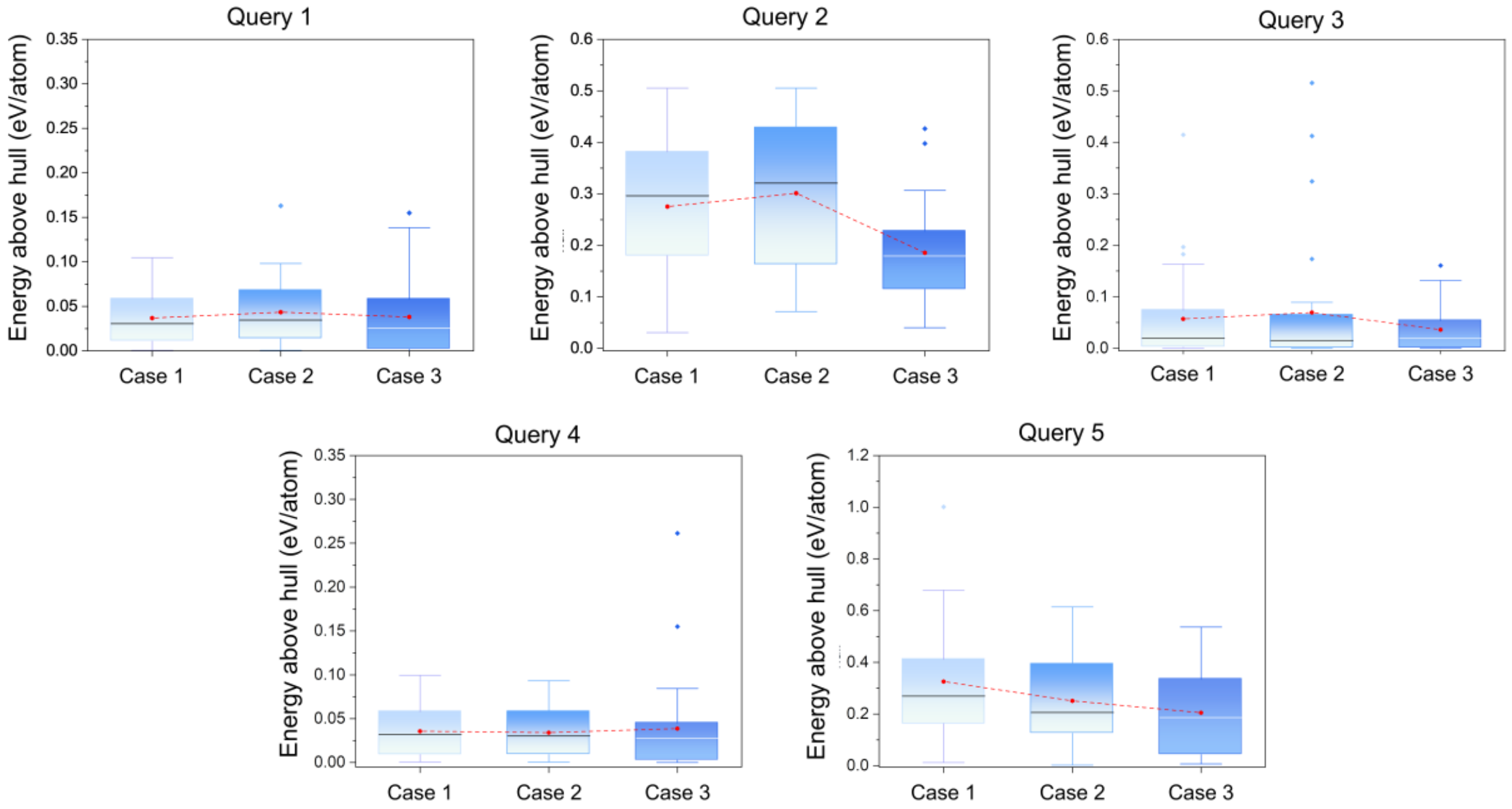

**Figure S2.** Results of energy above hull for generated DPs in cases 1-3 across all queries.

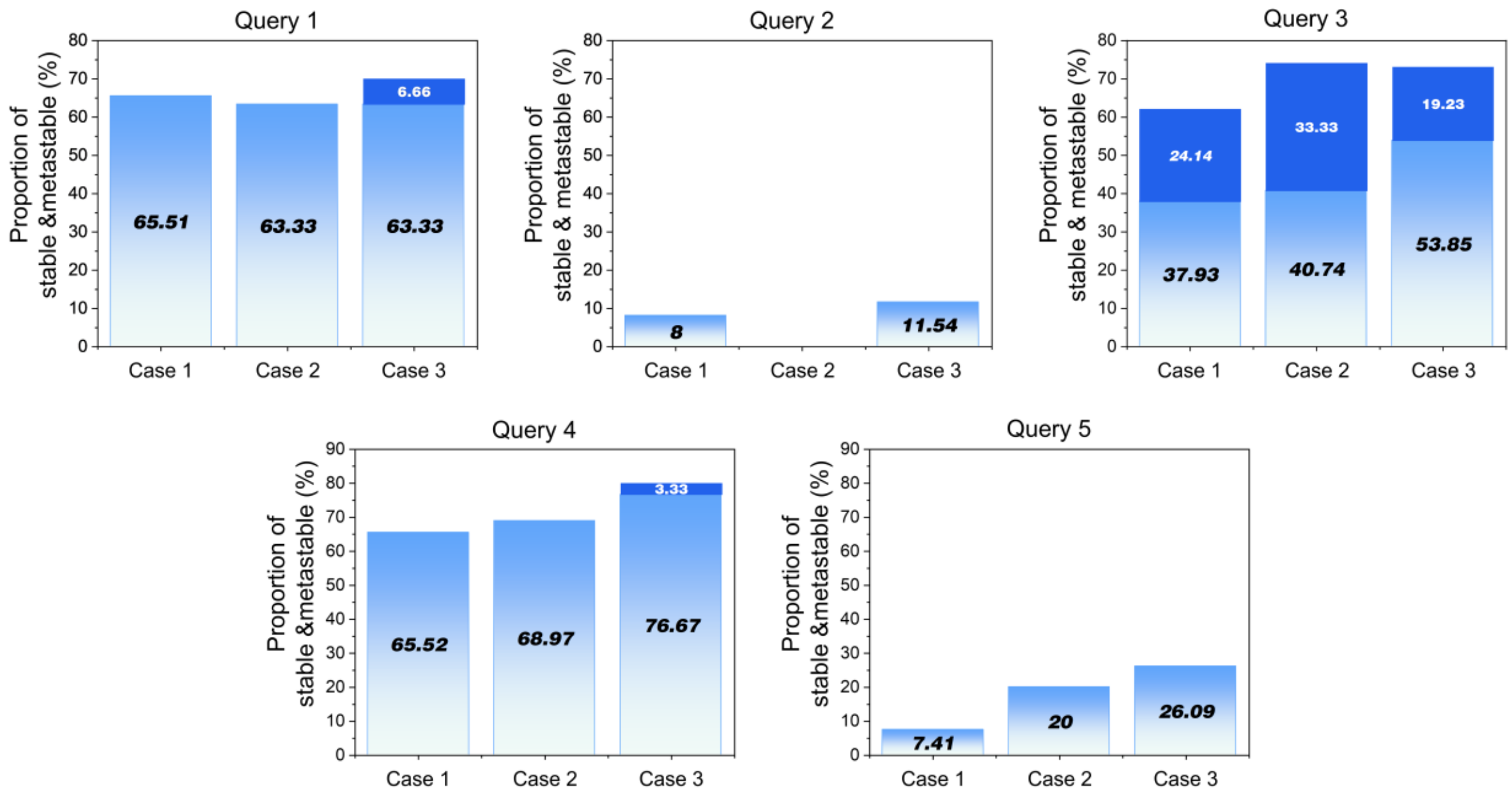

**Figure S3.** Fraction of generated DPs classified as stable or metastable states in cases1-3 across all queries.

## Supplementary Note 2. An example of violation scenario of the condition

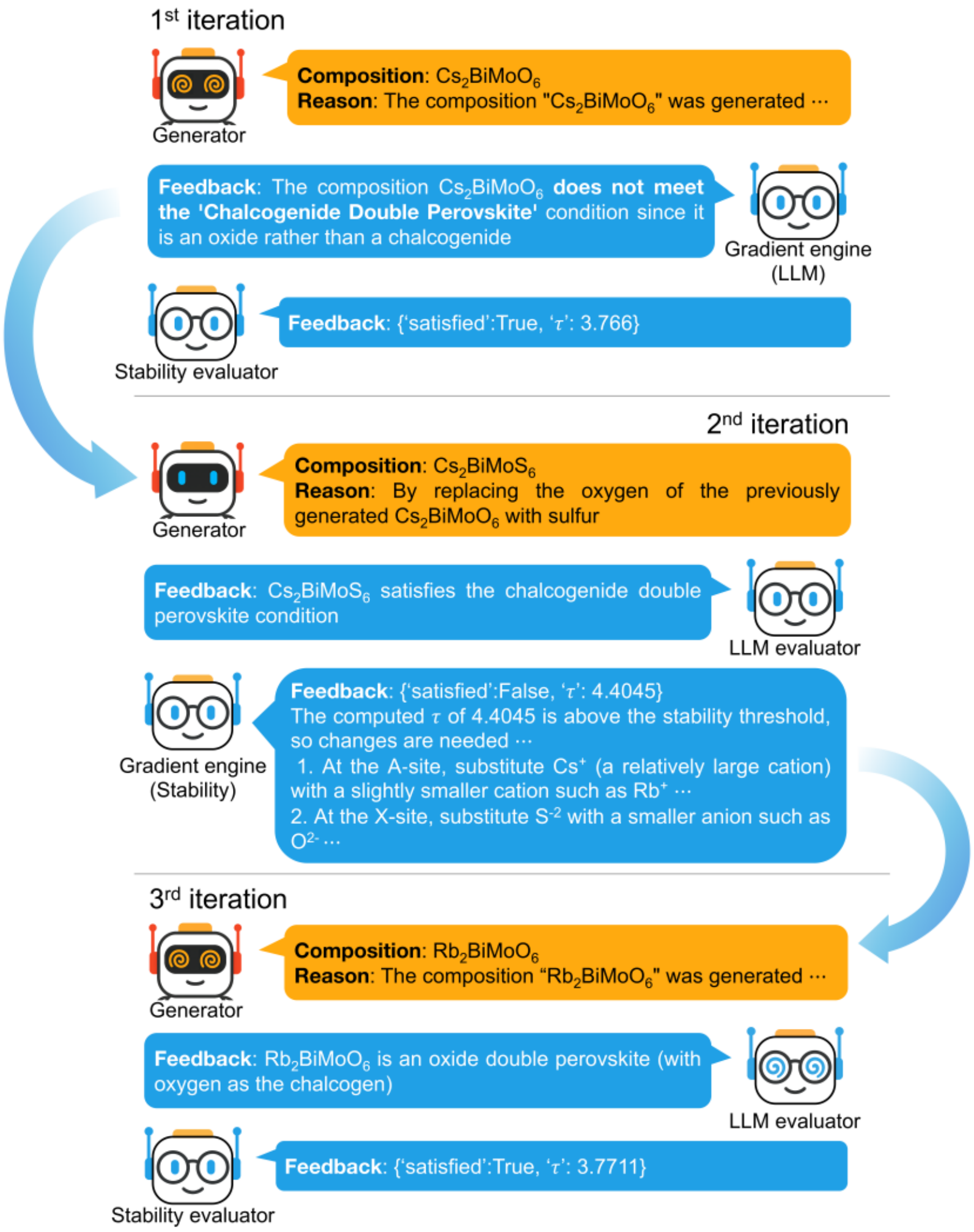

**Figure S4.** Hacking scenario emerging from iterative agent feedback. In the first iteration, the LLM evaluator effectively screens out formula proposals that do not satisfy user-specified conditions. In the second iteration, the *Knowledge-Informed Evaluator* (stability evaluator) incorrectly biases recommendations toward oxygen at the X-site. In the third iteration, the Formula Proposal Agents and LLM evaluator failed due to accumulated bias from previous recommendations, resulting in invalid outputs

**Supplementary Note 3. The number of rejections per each query**

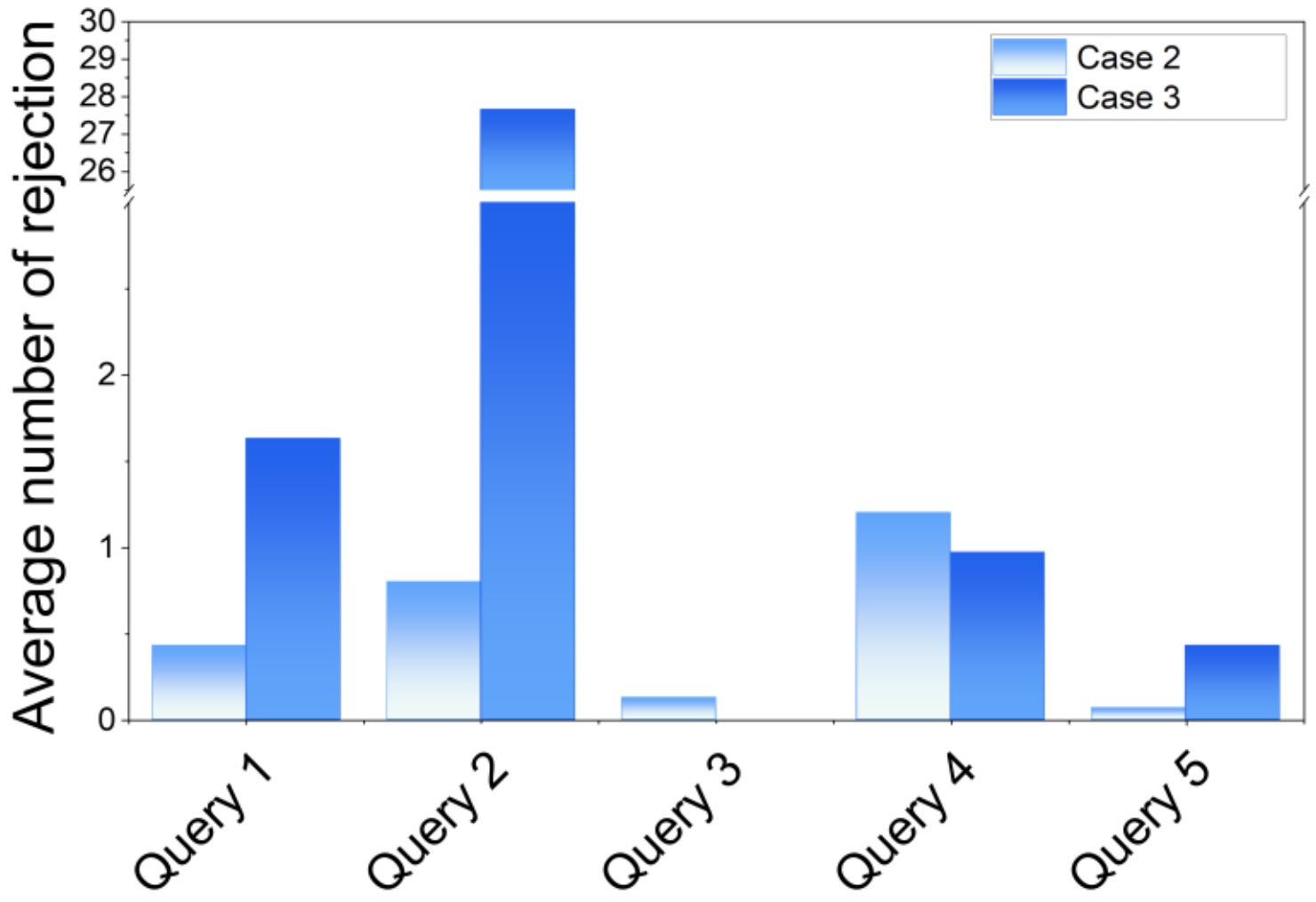

**Figure S5.** Average number of rejections per finally accepted composition for each query

**Supplementary Note 4. Knowledge probing for DP composition proposal**

Large language models (LLMs) are inherently black-box systems, and unless the developers explicitly disclose it, little information is available regarding the training data used in their development. Consequently, it is difficult to elucidate the patterns underlying their proposals for double perovskite (DP) composition by analyzing the training set. Recently, however, a growing body of research has explored techniques for eliciting the latent knowledge embedded within LLMs through carefully designed prompts.

In this section, we adopt such an approach to examine the knowledge tendencies of the o3-mini model in the context of DP composition proposal. Specifically, we prepared 30 paraphrased queries for each task of recommending stable halide and chalcogenide DPs. Each query was designed to elicit five compositions at a time, and to encourage diversity, each query was executed five times. As a result, in the first round, a total of 750 compositions were generated for halide and chalcogenide DPs respectively. Subsequently, in the second round, we prompted the model to propose alternative compositions for each query, excluding the five compositions that had already been generated.

For halide DPs, the o3-mini model exhibited a preference for $Cl^-$ and $Br^-$ (**Table S1**). In Round 1, it proposed 438 $Cl^-$-containing and 304 $Br^-$-containing compositions, whereas only 8 $I^-$-based candidates were identified. In Round 2, the number of $Cl^-$ and $Br^-$-containing compositions slightly decreased to 381 and 291, respectively, while $I^-$-containing compositions increased markedly to 78. This trend indicates that the model's knowledge is primarily aligned with $Cl^-$ and $Br^-$-based systems, and that the emergence of $I^-$-based compositions requires repeated exploration.

These results highlight that multiple iterations are essential for achieving diversity in halide DP proposals; without the generated space would remain dominated by $Cl^-$ and $Br^-$.

**Table S1.** Results of Round 1 and Round 2 for halide DP composition proposals.

| X site anion | Round 1 | Round 2 |
|---|---|---|
| $Cl^-$ | 438 | 381 |
| $Br^-$ | 304 | 291 |
| $I^-$ | 8 | 78 |

For chalcogenide DPs, the o3-mini model showed a strong preference for $S^{2-}$ across both rounds (**Table S2**). In Round 1, 602 $S^{2-}$-containing compositions were proposed, compared to 123 with $Se^{2-}$ and none with $Te^{2-}$. Interestingly, a small number of halide-based compositions, including 18 with $Cl^-$ and 7 with $Br^-$, also appeared despite the chalcogenide-specific prompt. In Round 2, $S^{2-}$-based compositions slightly decreased to 553, while $Se^{2-}$-based proposals increased to 192, and $Te^{2-}$-containing compositions emerged for the first time (n = 5). No halide-containing candidates were generated in this round. These results suggest that the model's knowledge is heavily biased toward $S^{2-}$, with $Se^{2-}$ explored to a lesser extent, and $Te^{2-}$ only marginally represented. Importantly, the disappearance of unintended halide compositions in Round 2 demonstrates that iterative prompting not only enhances diversity (e.g., by including $Te^{2-}$) but also improves adherence to the intended task conditions. Without such iterations, the generated space would have remained dominated by $S^{2-}$ with limited diversity and occasional off-target outputs.

**Table S2.** Results of Round 1 and Round 2 for halide DP composition proposals.

| X site anion | Round 1 | Round 2 |
|---|---|---|
| $S^{2-}$ | 602 | 553 |
| $Se^{2-}$ | 123 | 192 |
| $Te^{2-}$ | - | 5 |
| $Cl^{-}$ | 18 | - |
| $Br^{-}$ | 7 | - |

To further assess the impact of iterative prompting on compositional diversity, we examined the number of unique compositions generated among the 750 candidates for both halide and chalcogenide DPs (**Table S3**). The results show that the number of unique halide DP compositions increased from 16 in Round 1 to 71 in Round 2, while the number of unique chalcogenide DP compositions increased from 171 to 250. These findings provide additional evidence that our iterative approach to prompting enhances the diversity of generated DP compositions.

**Table S3**. The number of unique halide and chalcogenide DP compositions generated across rounds

| The number of unique compositions | Round 1 | Round 2 |
|---|---|---|
| Halide DP | 16 | 71 |
| Chalcogenide DP | 171 | 250 |

**Supplementary Note 5. Distribution of each site elements per query for Cases 1-3.**

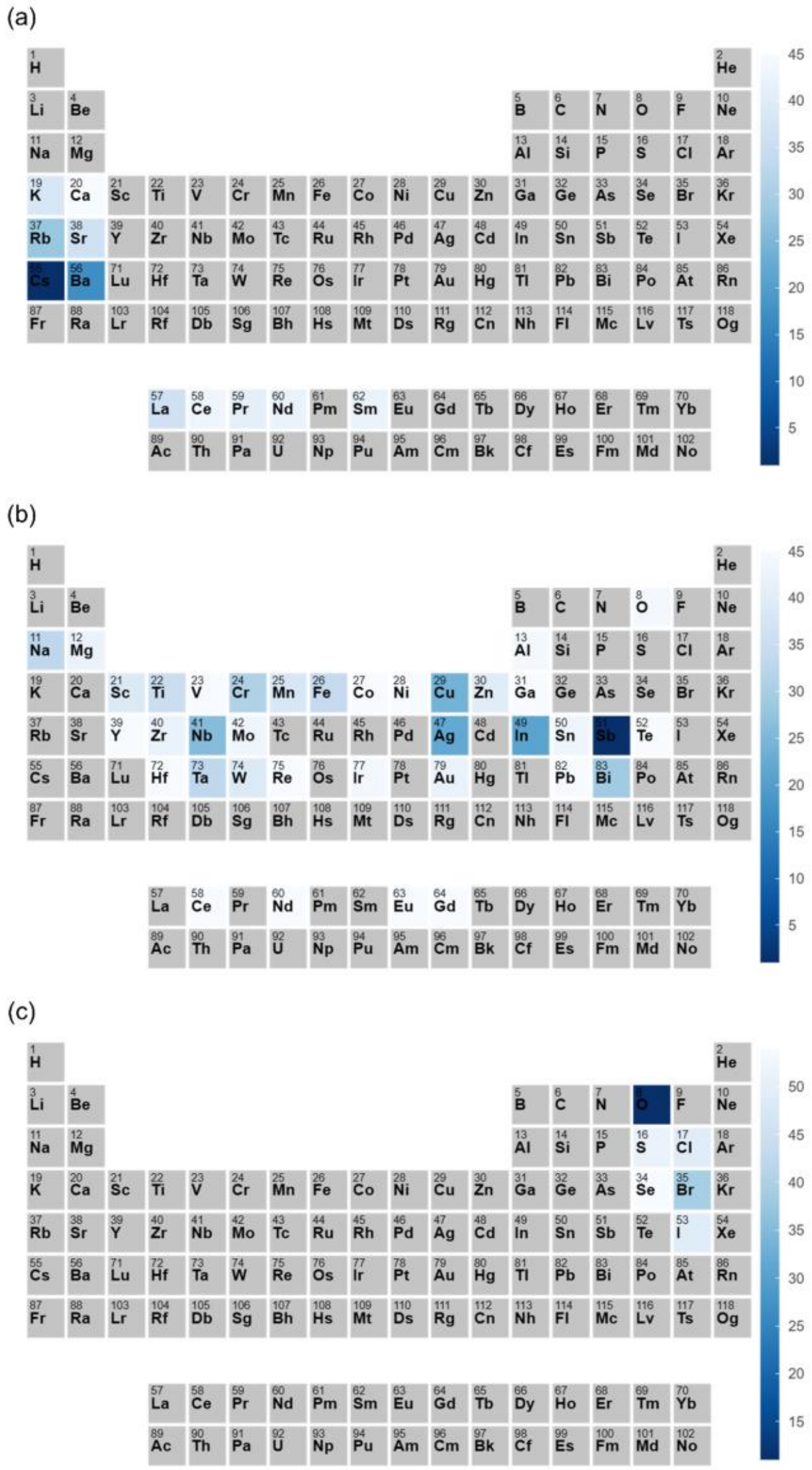

**Figure S6. (a)** Distribution of X-site ions of finally accepted double perovskite compositions across all queries for Case 1. **(b)** Distribution of B- and B'-site ions of finally accepted double perovskite compositions across all queries for Case 1. **(c)** Distribution of X-site ions of finally accepted double perovskite compositions across all queries for Case 1.

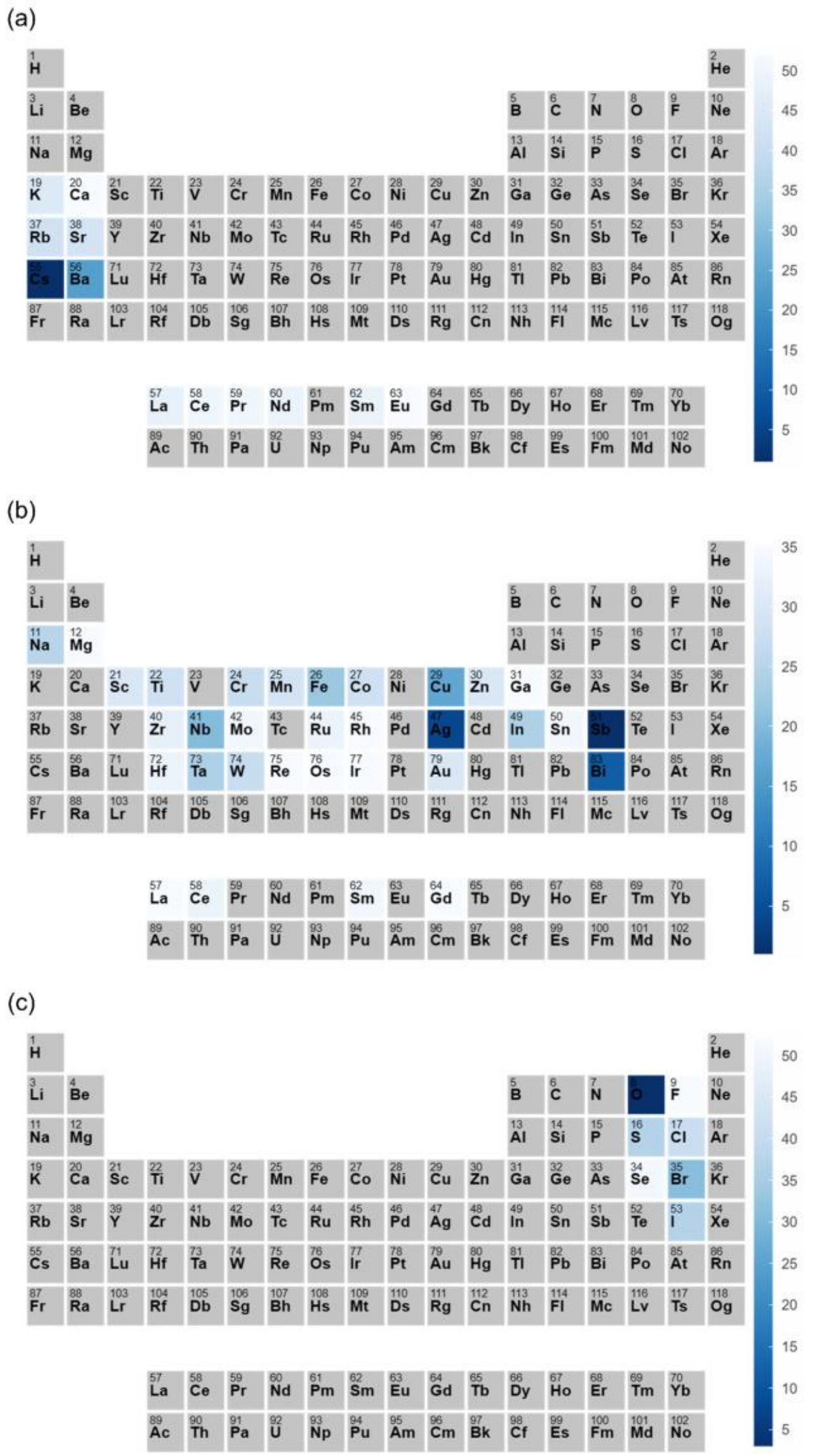

**Figure S7. (a)** Distribution of X-site ions of finally accepted double perovskite compositions across all queries for Case 2. **(b)** Distribution of B- and B'-site ions of finally accepted double perovskite compositions across all queries for Case 2. **(c)** Distribution of X-site ions of finally accepted double perovskite compositions across all queries for Case 2.

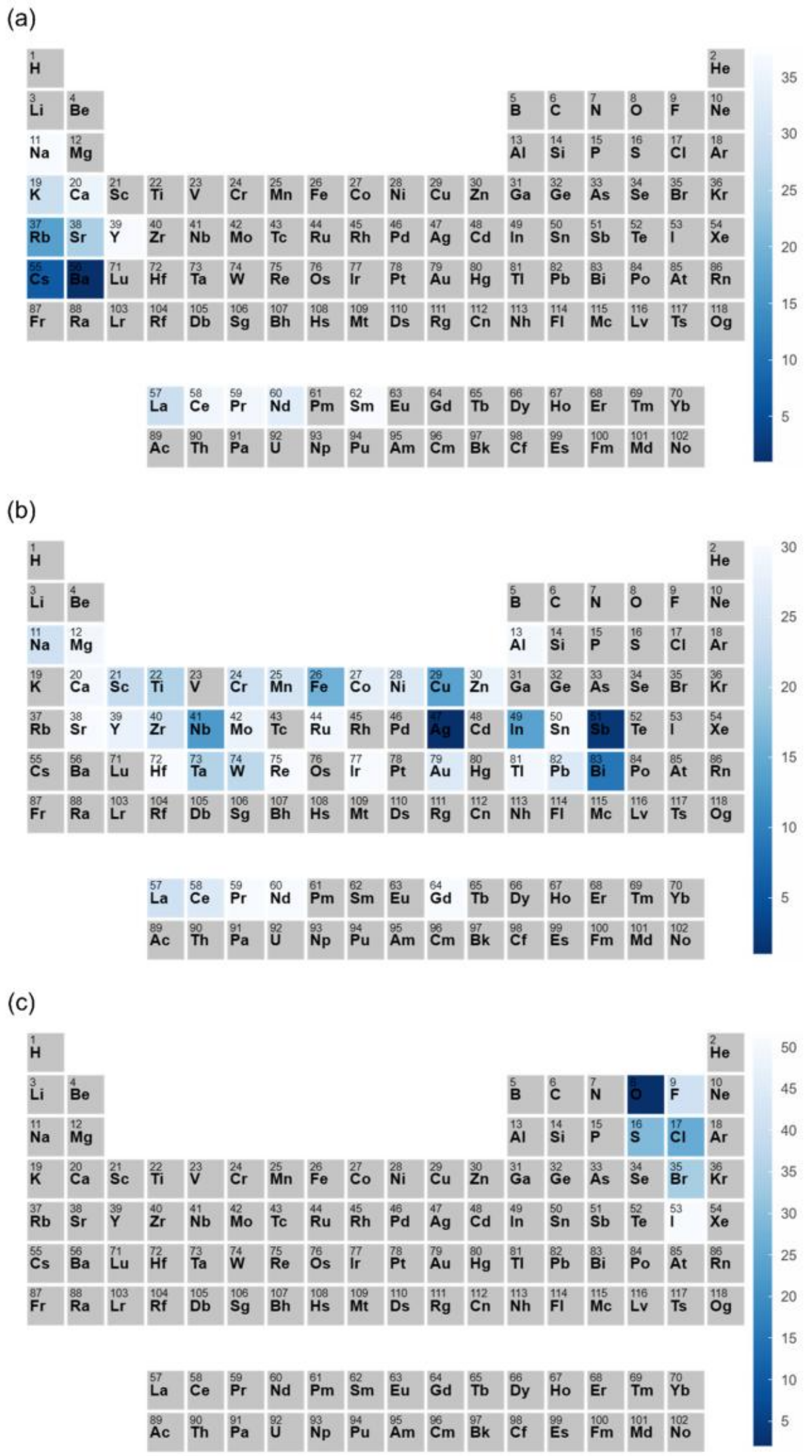

**Figure S8. (a)** Distribution of X-site ions of finally accepted double perovskite compositions across all queries for Case 3. **(b)** Distribution of B- and B'-site ions of finally accepted double perovskite compositions across all queries for Case 3. **(c)** Distribution of X-site ions of finally accepted double perovskite compositions across all queries for Case 3

## Supplementary Note 6. Predictive performance metrics of CrabNet

**Table S4**. Classification performance of the CrabNet for MP data.

| Query | Macro average | | | Weighted average | | |
|---|---|---|---|---|---|---|
| | Precision | Recall | F1-score | Precision | Recall | F1-score |
| **1** | 0.93 | 0.76 | 0.81 | 0.90 | 0.88 | 0.86 |
| **2** | 0.53 | 0.57 | 0.31 | 0.86 | 0.33 | 0.40 |
| **3** | 0.80 | 0.73 | 0.75 | 0.83 | 0.84 | 0.83 |
| **4** | 0.94 | 0.74 | 0.80 | 0.91 | 0.90 | 0.88 |
| **5** | 0.51 | 0.53 | 0.42 | 0.78 | 0.49 | 0.56 |